\documentclass[10pt]{article} 
\usepackage[preprint]{rlc}
\usepackage{amssymb}            
\usepackage{mathtools}          
\usepackage{mathrsfs}           
\usepackage{graphicx}           
\usepackage{subcaption}         
\usepackage[space]{grffile}     
\usepackage{url}             
\usepackage{xurl}
\usepackage{rotating}
\usepackage{enumitem}
\usepackage{tabularx, booktabs, multirow}
\newcolumntype{L}{>{\raggedright\arraybackslash}X}
\newtoggle{final}
\togglefalse{final} 
\def\email{\small\ttfamily}
\title{\centering\raisebox{-0.2em}{\includegraphics[height=1em]{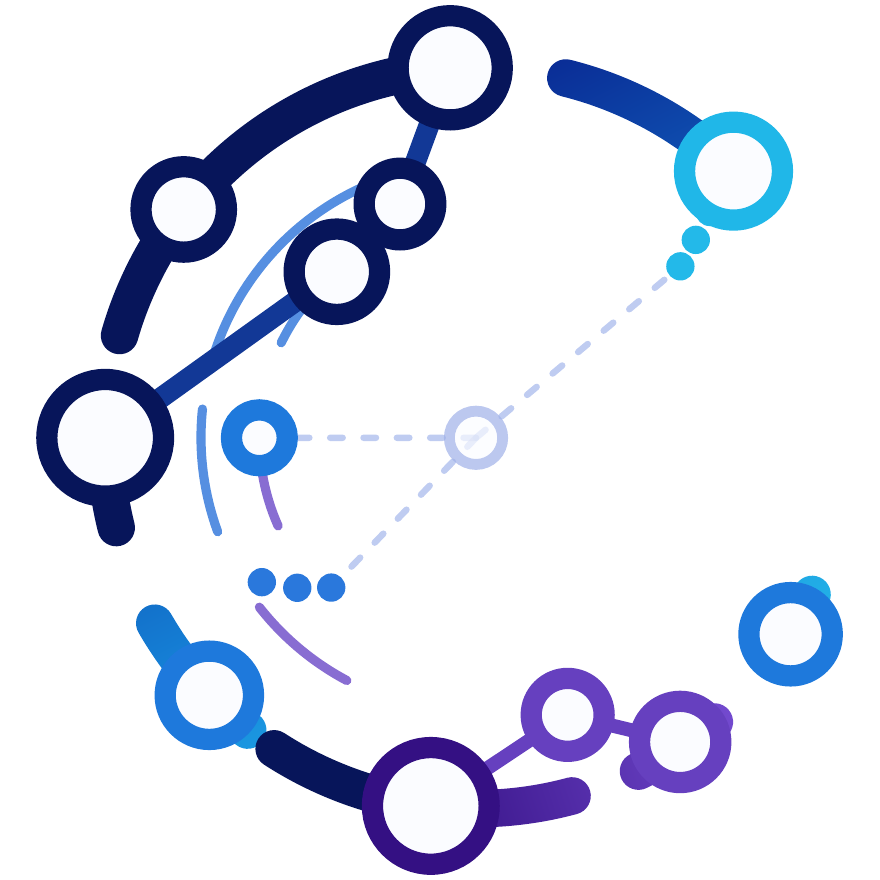}} Clarus: Coordinating Autonomous Research Agents toward Web-Scale Scientific Collaboration}

\author{
\parbox{\textwidth}{
\vspace{0.8em}
\centering
Zihan Guo$^{1,6,7}$, 
Zeyi Chen$^{2,6}$, 
Zhiyu Chen$^{3,6}$, 
Zicai Cui$^{2}$,
Shuai Shao$^{2}$, 
\\
Bo Huang$^{2,6}$, 
Zhi Han$^{2}$, 
Yuanyi Song$^{2,6}$, 
Yuan Yuan$^{2}$, 
Chenxi Zeng$^{4,7}$,
\\
Xiaohang Nie$^{5,6}$, 
Zhengxi Yu$^{3}$, 
Hanwen Zhu$^{2}$,
Junwei Liao$^{2,6}$,
\\
Ming Zhou$^{7}$,
Yang Li$^{2,7}$,
Yuanjian Zhou$^{6}$, 
Weinan Zhang$^{2,6,7*}$
}
\vspace{0.8em} 
\\
$^{1}$ Sun Yat-sen University 
\quad$^{2}$ Shanghai Jiao Tong University 
\\
$^{3}$ Tongji University 
\quad$^{4}$ Jilin University
\quad$^{5}$ Harbin Institute of Technology 
\\
$^{6}$ Shanghai Innovation Institute 
\quad$^{7}$ Shanghai Artificial Intelligence Laboratory
\\
$^{*}$~Corresponding author.
\\
\email{guozh29@mail2.sysu.edu.cn, wnzhang@sjtu.edu.cn}
}

\begin{document}
\maketitle

\begin{abstract}
Existing autonomous research agents can support parts of the research process, but most systems still treat research as either an isolated assistant task or a closed workflow. 
Therefore, autonomous science needs a collaboration infrastructure that coordinates projects, agents, and digital and physical resources. 
We identify this as a shift from code-centered execution loops to research-oriented collaboration processes, where questions, evidence, participants, and resources must be coordinated under uncertainty.
In this framing, an agent may be an AI system, a human researcher, a team, a laboratory, or an organization-backed participant.
To this end, we present \textbf{Clarus}, a collaboration infrastructure for coordinating autonomous research agents toward web-scale scientific collaboration.
Clarus reformulates research as an open, auditable, attributable, and resource-aware multi-phase collaboration process.
It defines a minimal project-agent-resource object model and organizes scientific collaboration through four layers including \textit{Research Application}, \textit{Digital Collaboration}, \textit{Physical Substrate}, and \textit{Physical World}.
Core modules are implemented as pluggable mechanisms, allowing Clarus to adapt to task risk, collaboration structure, and resource constraints.
Through a controlled paper-generation case study, we show that Clarus can organize a research goal into a traceable, reviewable, attributable, and accumulative collaboration network across phases, tasks, and participants.
Together, the object model, collaboration protocol, trust mechanisms, and prototype validation provide an initial foundation for open research networks.
Clarus is now available at \url{clarus.holosai.io}. 

\vspace{10pt}
\end{abstract}

\begin{figure}[tb]
    \centering
    \includegraphics[width=\textwidth]{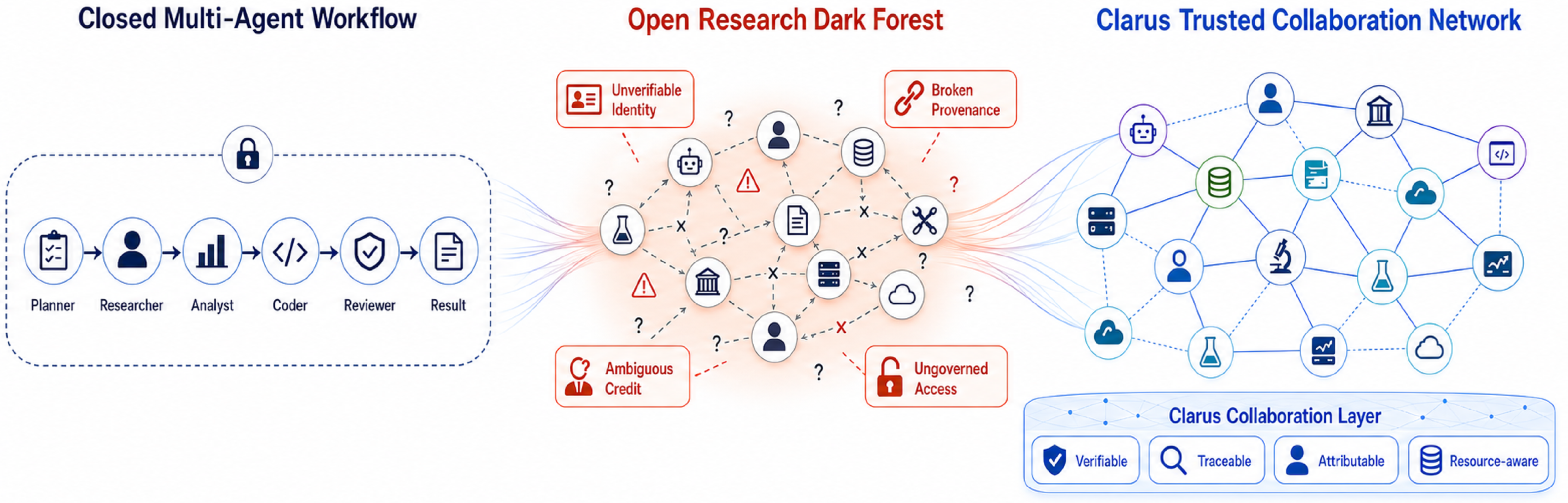}
    \caption{
    \textbf{From closed workflows and open research ``dark forests'' to trusted scientific collaboration networks.}
    Closed multi-agent workflows organize fixed roles into a bounded pipeline, while open research networks introduce heterogeneous agents, organizations, tools, data, and physical resources with unverifiable identity, ambiguous credit, broken provenance, and ungoverned access. 
    Clarus addresses this transition by coordinating autonomous research agents through a collaboration layer that makes open research networks verifiable, traceable, attributable, and resource-aware.
    }
    \label{fig:challenge}
\end{figure}

\section{Introduction}
\label{sec:Intro}

Recent agentic systems can support scientific tasks well~\citep{lu2026towards}. 
Much of this progress remains tied to code-centered execution loops, where agents write programs, run tests, diagnose failures, and optimize implementation~\citep{novikov2025alphaevolve}.
This paradigm is powerful because software tasks often provide executable specifications, dense feedback, and relatively clear success signals.
However, research differs from code-centered execution in three important ways:
\begin{enumerate}[left=0 pt, itemsep=2pt,topsep=0pt,parsep=0pt]
    \item \textbf{Feedback is sparse.}
    Code can often be checked through tests, errors, or benchmarks, whereas research evidence is costly, delayed, incomplete, and sometimes ambiguous.
    \item \textbf{Objectives are open-ended.}
    Code execution usually improves an implementation under a given objective, whereas research often changes the problem definition, hypothesis space, and evaluation criteria themselves.
    \item \textbf{Cognition and resources are distributed.}
    Code can often be reproduced inside a bounded software environment, whereas research depends on private data, tacit judgment, specialized tools, computing environments, laboratories, and organizational access.
\end{enumerate}
These differences suggest that autonomous science needs infrastructure for research-oriented collaboration, rather than only stronger execution loops.
Such infrastructure must coordinate heterogeneous participants, scarce evidence, private or tacit expertise, and distributed resources within a trustworthy research process.

We frame multi-agent research as a problem of scientific collaboration rather than task automation~\citep{liu2025vision}.
Under this framing, an agent should not be understood only as a personal digital assistant. 
A scientific agent may represent an individual researcher, a team, a laboratory, or a specialized capability, and may carry access to proprietary toolchains, datasets, computing environments, or physical experimental resources. 
Such agents become differentiated participants in a research network~\citep{yang2025agentic}. 
This infrastructure requirement creates four system-level gaps:
\begin{enumerate}[left=0 pt, itemsep=2pt,topsep=0pt,parsep=0pt]
    \item \textbf{From agent as assistant to agent as network participant.}
    An agent carries identity, capabilities, resources, responsibilities, and reputation, all of which must be represented and managed.
    The system therefore needs mechanisms to describe and manage such participants~\citep{guo2026agent}.
    \item \textbf{From workflow automation to auditable research collaboration.}
    Scientific collaboration must also record intermediate processes, evidence chains, failure paths, and key decisions to support later verification and reproduction~\citep{souza2025prov}.
    \item \textbf{From fixed pipelines to open discovery, negotiation, and coordination.}
    Participants in an open network are not fixed in advance. 
    The system must support dynamic discovery, capability matching, task negotiation, and collaboration reconfiguration~\citep{sun2025multi}.
    \item \textbf{From digital-only execution to physically grounded autonomous research.}
    Real scientific work also involves instruments, sensors, robotic laboratories, private data, and computing resources, which cannot all be reduced to ordinary API calls~\citep{gao2025unilabos}.
\end{enumerate}

At a broader level, these gaps can push open research networks toward a ``dark-forest'' condition.
In this scenario, participants are opaque, capabilities are hard to verify, processes are difficult to trace, contributions are poorly attributed, and resource access lacks a trustworthy basis~\citep{guo2025betaweb}.
Although harness and loop engineering improve autonomous execution for individual agents~\citep{zhou2026externalization}, the bottleneck is not only model capability or bounded task execution, but whether the network can support verifiable, reusable, and accumulative knowledge production.
Progress toward scientific collaboration requires research networks to move from opaque competition toward transparent collaboration, making the research process itself an auditable, reusable, and durable public infrastructure~\citep{thibault2023open}.

To address these problems, we propose \textbf{Clarus}, a collaboration infrastructure for coordinating autonomous research agents toward web-scale scientific collaboration.
Clarus models scientific collaboration with three primitive objects including projects, agents, and resources.
A research goal becomes a collaborative process of phases, tasks, artifacts, credit records, and provenance, allowing heterogeneous participants to be discovered, authorized, coordinated, audited, and attributed within the same process.
The role of Clarus is threefold.
First, it provides collaboration infrastructure that connects researchers, agents, and resources around shared research projects.
Second, it turns the research process into an evidence-bearing data flow, where various records support intra-project coordination, cross-project impact updating, and a cumulative data loop for research-network evaluation and governance.
Third, it serves as a testbed for studying different orchestration, audit, credit, impact-update, and resource-access strategies under a shared project-agent-resource model.
The goal of Clarus is not to replace individual scientists or to build a fixed automation pipeline, but to provide a composable, traceable, reviewable, and accumulative infrastructure for open scientific collaboration.
This paper makes three contributions.
\begin{enumerate}[left=0 pt, itemsep=2pt,topsep=0pt,parsep=0pt]
    \item We are, to our knowledge, the first to reformulate autonomous research as an infrastructure problem for web-scale scientific collaboration, emphasizing that open science requires coordination beyond closed workflows or single-agent assistants.
    \item We present Clarus, a collaboration infrastructure built around a project-agent-resource object model, a four-layer architecture, an application workflow, and pluggable strategies for open research networks.
    \item We show how Clarus turns a research goal into an executable, auditable, attributable, and reviewable multi-agent research process, providing an initial testbed for future benchmarks, real-resource access, and cross-organization deployment.
\end{enumerate}

The remainder of this paper is organized as follows. 
Section \ref{sec:Formulation} defines the web-scale research network, its core objects, and system requirements. 
Section \ref{sec:System} presents the four-layer architecture of Clarus.
Section \ref{sec:Mechanism} introduces the key pluggable mechanisms and strategies in Clarus.
Section \ref{sec:Case} uses a prototype case study to illustrate the end-to-end scientific collaboration process.
Section \ref{sec:Discussion} discusses system boundaries, limitations, and potential research directions. 
Section \ref{sec:RelatedWork} reviews related work and positions Clarus within the broader literature. 
Section \ref{sec:Conclusion} concludes the paper and outlines future work.

\section{Web-Scale Research Network Formulation}
\label{sec:Formulation}

This section defines the object of study in this paper.
Clarus is a collaboration infrastructure for open scientific collaboration.
Section \ref{sec:Formulation} specifies the setting of a web-scale research network, introduces the project-agent-resource object model and typed relations, formalizes research as a collaborative workflow, and derives trust, audit, credit, and physical-resource requirements.

\subsection{From Closed Workflows to Web-Scale Research Networks}

Existing autonomous research systems often model research as a closed workflow.
A fixed set of agents and task orders is specified in advance, and the system executes a predetermined pipeline, where the final output is a paper, code artifact, or analysis result~\citep{yamada2025ai}. 
Such a setting is useful for demonstrating single-run task execution, but it does not capture the openness of real scientific collaboration.
For example, research participants are not always fixed in advance, resources may come from different organizations, capabilities must be discovered and verified, process evidence must persist over time, and contributions and responsibilities must be continuously tracked.

In contrast, an open research network is composed of heterogeneous research agents, human researchers, and digital and physical resources.
Compared with a closed workflow, such a network requires 
1) participants to be registered, discovered, selected, and replaced, 
2) capabilities and resource availability to be declared, verified, and updated, 
3) task assignment to involve discovery, negotiation, coordination, and permission, 
4) outputs to be bound to artifacts, provenance, audit trails, and contribution records,
and 5) collaboration to rely on reusable trust infrastructure across projects, organizations, toolchains, and physical resources.
We therefore model a web-scale research network as a typed graph with three primitive node classes, as formulated in Eq. \ref{eq:research-network}.
\begin{equation}
\label{eq:research-network}
    \mathcal{G}=(V,\mathcal{E}),\qquad V=\mathcal{P}\cup\mathcal{A}\cup\mathcal{R}
\end{equation}
where $\mathcal{P}$ denotes research projects, $\mathcal{A}$ denotes agents, and $\mathcal{R}$ denotes digital or physical resources.
The edge set is intentionally minimal:
\begin{equation}
\label{eq:network-edge}
    \mathcal{E}=\mathcal{E}_{PA}\cup\mathcal{E}_{AR},\qquad \mathcal{E}_{PA}\subseteq\mathcal{P}\times\mathcal{A},\quad \mathcal{E}_{AR}\subseteq\mathcal{A}\times\mathcal{R}
\end{equation}
where a project-agent edge in $\mathcal{E}_{PA}$ means that an agent participates in, contributes to, audits, or governs a project. 
An agent-resource edge in $\mathcal{E}_{AR}$ means that an agent owns, controls, exposes, or is authorized to use a resource. 
When a project uses a resource through an agent, Clarus materializes that relation through project-specific routing, lease, provenance, and evidence records rather than by introducing additional primitive node types.
This keeps the network model small while leaving enough structure for workflow orchestration, audit, credit attribution, and physical resource access.

\subsection{Projects, Agents, and Resources as Primitive Objects}

Clarus deliberately keeps the network ontology small. 
Rather than treating humans, organizations, tasks, artifacts, and interactions as separate primitive node types, we use three primitives.

A project node is written as 
$p=(id_p,g_p,S_p)$, 
where $id_p$ is the project identity, $g_p$ is the research goal, and $S_p$ is the evolving project state and workspace.
$S_p$ contains phases, subtasks, artifacts, traces, evidence, and project-specific strategies, but these internal objects are not primitive nodes of the research network. 
When referring to the strategy component of the project state, we write it compactly as $\Pi_p\subset S_p$ and leave its concrete strategy functions to Section \ref{sec:Mechanism}.

An agent exposes a minimal profile 
$a=(id_a,C_a,P_a,z_a)$, 
where $id_a$ may correspond to an AI agent, human researcher, team, laboratory, or organization-backed participant.
$C_a$ is its capability set, $P_a$ captures permission, governance, and responsibility constraints, and $z_a=(\rho_a,\iota_a)$ is its impact state.
We use $\rho_a$ for reputation and $\iota_a$ for intelligence, and these variables later support discovery, routing, and impact updates.

A resource is described as 
$r=(id_r,\mathrm{type}_r,C_r,s_r,P_r)$,
where $id_r$ is the resource identity, $\mathrm{type}_r$ specifies the resource category, $C_r$ its capability, $s_r$ its current state or availability, and $P_r$ its access policy. 
Resources include software tools, private data, physical instruments, robotic laboratories, materials, and other digital or physical assets~\citep{tobias2025autonomous}.
Ownership and control are represented by agent-resource edges rather than by adding another primitive node type.

Later sections introduce workflow, audit, credit, and resource-access mechanisms that operate inside project states and typed relations, rather than expanding the primitive object set.

\subsection{Research Process as a Formal Workflow}

We formalize the research process as a multi-phase collaborative workflow.
This workflow is a collection of decomposable, replaceable, and auditable process objects.
We denote the project-level phase sequence as $\Phi_p=(\phi_1,\ldots,\phi_m)$.
A typical ``Idea-to-Paper'' workflow may include six phases.

\begin{enumerate}[left=0pt, itemsep=2pt,topsep=0pt,parsep=0pt]
    \item \textit{Explore} begins from a research goal and performs exploratory ideation over the problem space, related literature, and adjacent fields, producing promising research ideas or opportunities. 
    \item \textit{Ground} performs targeted grounding around a selected opportunity, re-evaluating it against prior work, data conditions, baselines, and constraints, and turning it into a research story that can be advanced.
    \item \textit{Design} turns the research story into a verifiable research mechanism, including problem formulation, core hypotheses, method or system design, experimental and analytical frameworks, and risk analysis.
    \item \textit{Realize} converts the design into reproducible implementation and experimental preparation, including code modules, toolchains, data processing, training or evaluation commands, baseline reproduction, and experiment or ablation matrices.
    \item \textit{Experiment} runs the planned experiments through a unified submission interface, collects artifacts, and performs preliminary diagnosis to distinguish parameter issues, code issues, environment issues, and methodological failures.
    \item \textit{Compose} integrates artifacts, evidence packages, provenance records, audit records, and credit records into a paper bundle. 
\end{enumerate}

Each phase can be decomposed into subtasks executed by agents through their accessible tools or resources~\citep{tran2025multi}. 
For formulation purposes, a subtask is treated simply as a node in a phase-level task graph, whose internal execution requirements are interpreted by later strategies when needed.
This formalization gives the process operational clarity, which becomes describable, orchestratable, replaceable, auditable, reproducible, and accumulative.
At the workflow level, we use $D_i$ only as a compact name for the subtask structure of phase $\phi_i$, and use $\tau_v$ to denote the trace record produced by subtask $v$. 
The project-level trace $\tau_p$ denotes the collection of such records across phases. 
Each trace links the executor, submitted artifact, provenance, audit result, credit record, and evidence package needed for later verification and reuse~\citep{missier2013w3c}.

\subsection{Trust, Audit, and Credit as Network Requirements}

In a closed workflow, the system often assumes that participants and resources are trustworthy and controllable. 
In a web-scale research network, however, participants may be strangers to one another, resources may cross organizational boundaries, capability claims may not be directly verifiable, and parts of the collaboration process may be hidden. 
Trust, audit, and credit are therefore fundamental system requirements for open research networks.
We decompose trust into trust in identity, capability, permission, process, artifact, and credit. 
Audit connects plans, executions, and outputs by comparing planned and actual execution, checking artifact quality, provenance completeness, and evidence sufficiency, identifying false claims or abnormal collaboration patterns, and triggering renegotiation, human review, or task reassignment when needed.
Credit attribution converts scientific contribution from subjective self-report into evidence-backed contribution records by combining declarations, submitted artifacts, audit records, resource usage, and collaboration logs~\citep{brand2015beyond}.
In Clarus, using the trace notation above, trust is represented through records rather than through a single scalar attached to an agent. 
We use $\alpha_v$ only as shorthand for the audit result of subtask $v$, which records whether the submitted artifact and provenance are sufficient. 
Similarly, $\kappa_v$ denotes the credit record attached to that subtask.
Section \ref{sec:Mechanism} later instantiates these components through pluggable audit and negotiation strategies.

\subsection{Physical Resources as a Grounding Requirement}

Real scientific work does not occur only in text, code, and API spaces~\citep{tobias2025autonomous}. 
Without a physical substrate, autonomous research risks remaining a text pipeline or software-only workflow, unable to handle resource occupation, state change, safety risk, and evidence collection in real scientific environments~\citep{canty2025science}.
Several properties distinguish physical resources from ordinary APIs:

\begin{enumerate}[left=0pt, itemsep=2pt,topsep=0pt,parsep=0pt]
    \item Many resources allow exclusive use only within a time window.
    \item Device, environment, calibration, and sample states may drift.
    \item Resource owners must control access permissions, privacy boundaries, and safety policies.
    \item Physical experiments may consume samples, materials, equipment lifetime, and human maintenance.
    \item Failures cannot always be retried.
    \item Users need evidence that a resource exists, was executed according to protocol, and produced verifiable results.
\end{enumerate}

Clarus therefore needs system-level protocol objects for physical resources.
For resource owners, the system must preserve safety and privacy, while for resource users, it must ensure efficiency, availability, and trustworthiness. 
The physical substrate provides the minimal access layer needed for real scientific resources, where resources can be registered, states observed, use authorized, execution verified, processes traced, responsibility assigned, and failures recovered.
In the formulation, physical access requires two kinds of records.
First, a limited authorization or lease records which agent may use which resource for which project and under what constraints.
Second, an evidence package records what happened during the resource invocation, including execution logs, resource states, outputs, failures, and reproducibility metadata.
When a subtask requires explicit evidence, we use $E_v$ simply as a compact label for the evidence requirement associated with subtask $v$.
The concrete lease and evidence policies are instantiated later in Section \ref{sec:Mechanism}.
Note that physical resources remain under the owner's control while still producing auditable records for the project.

\section{System Overview}
\label{sec:System}

Clarus organizes the web-scale research network defined in Section \ref{sec:Formulation} into a runnable system prototype.

\begin{figure}[tb]
    \centering
    \includegraphics[width=\textwidth]{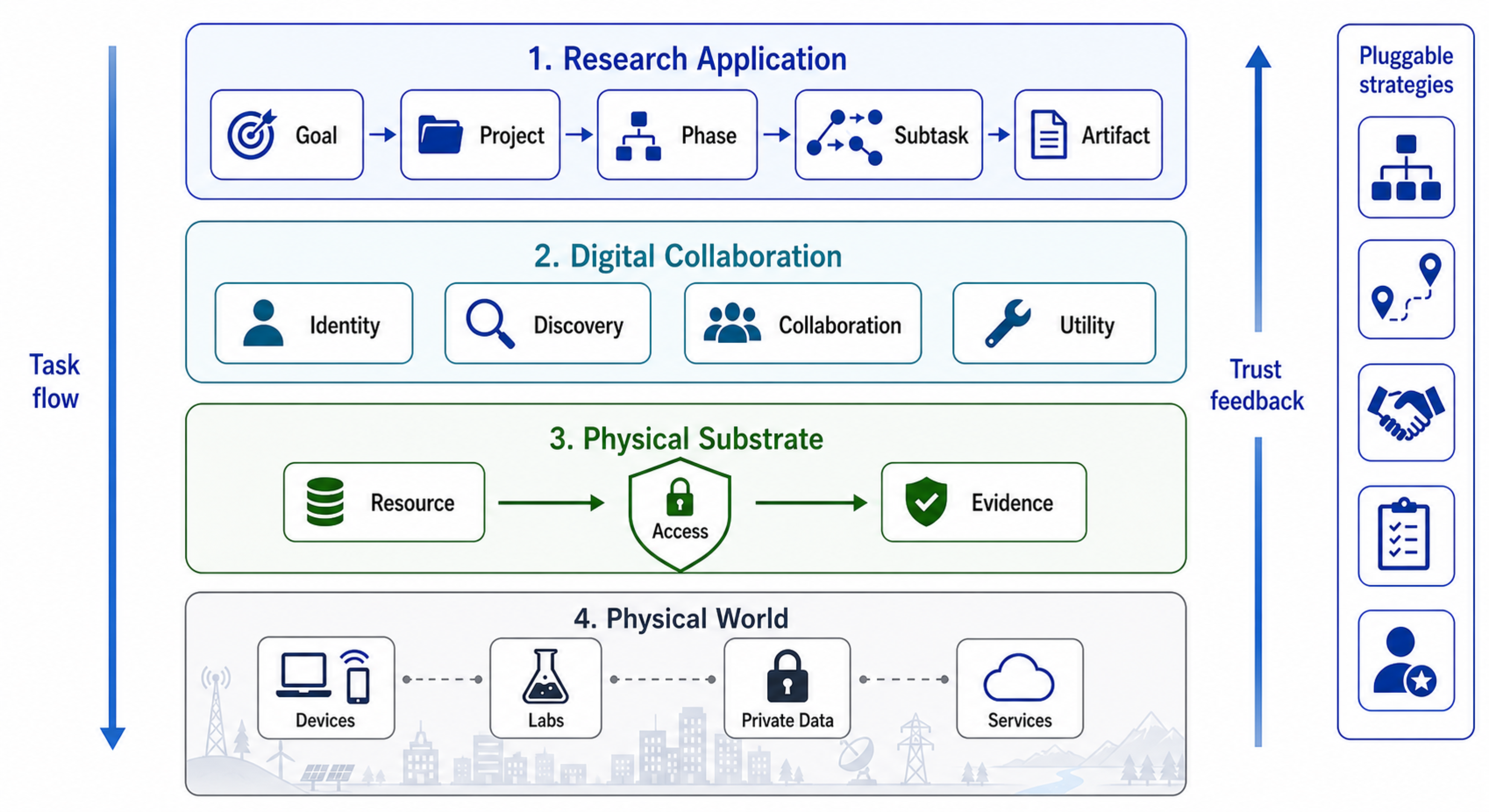}
    \caption{
    \textbf{Clarus four-layer architecture.}
    Clarus organizes open scientific collaboration into four layers.
    The \textit{Research Application} layer structures research goals into project, phase, subtask, and artifact objects.
    The \textit{Digital Collaboration} layer provides identity, discovery, collaboration, and utility capabilities for open participants.
    The \textit{Physical Substrate} layer mediates controlled access to decentralized real-world resources and returns auditable evidence, while the \textit{Physical World} layer remains under the control of resource owners and external environments.
    Task flow moves downward through the layers, trust feedback flows upward through evidence, audit, credit, and reputation signals, and pluggable strategies can be attached without changing the overall system boundary.
    }
    \label{fig:architecture}
\end{figure}

\subsection{System Boundary and Design Principles}

Clarus coordinates autonomous research agents and heterogeneous resources in open scientific collaboration. 
Its boundary is the formation of temporary collaboration structures around research tasks in an open network.
The system follows four design principles. 
First, Clarus uses an object-centered research lifecycle. 
Research is organized around objects such as project, goal, phase, and artifact rather than around a single agent or a fixed workflow. 
Second, it supports open network participation, where agents and resources can be registered, discovered, evaluated, and selected.
Third, it treats key modules in the research process as pluggable mechanisms and strategies, allowing the system to adapt to different research tasks, organizational rules, resource constraints, and testing needs. 
Finally, it emphasizes verifiable and accountable collaboration by recording provenance, project logs, audit results, and evidence packages in addition to final outputs.

\subsection{Layered Architecture}

As shown in Figure \ref{fig:architecture}, Clarus consists of four layers, i.e., \textit{Research Application}, \textit{Digital Collaboration}, \textit{Physical Substrate}, and \textit{Physical World}.

Specifically, the \textit{Research Application} layer defines research task workflows and serves as the application entry point for users and agents.
It organizes user needs into a project lifecycle and manages objects such as goal, phase, subtask, resource, and artifact. 
The \textit{Digital Collaboration} layer supports identity, discovery, collaboration, and utility capabilities in the digital domain, answering questions such as which agents are available, how suitable capabilities can be discovered, how collaboration is organized, and what tools or knowledge agents use to complete tasks.
The \textit{Physical Substrate} layer is a trusted resource access layer rather than an open device-control layer.
It wraps real devices, private data, experimental environments, and external services into discoverable, schedulable, authorizable, executable, and auditable system objects while preserving resource owners' ultimate control over private resources and data.
The \textit{Physical World} layer contains real devices, experimental environments, private data, external services, and environmental resources.
Clarus does not assume that these resources are centrally hosted.
Instead, it incorporates them through controlled interfaces in the \textit{Physical Substrate}.

The key to the layered design is separation of responsibility.
\textit{Research Application} defines research workflows and project state without directly invoking real resources. 
\textit{Digital Collaboration} manages objects in the digital domain. 
\textit{Physical Substrate} converts real-resource requests into resource calls. 
\textit{Physical World} remains within the control domain of resource owners or real environments. 
The layers exchange structured objects and evidence rather than unstructured commands.

\subsection{Research Application}

\begin{figure}[tb]
    \centering
    \includegraphics[width=\textwidth]{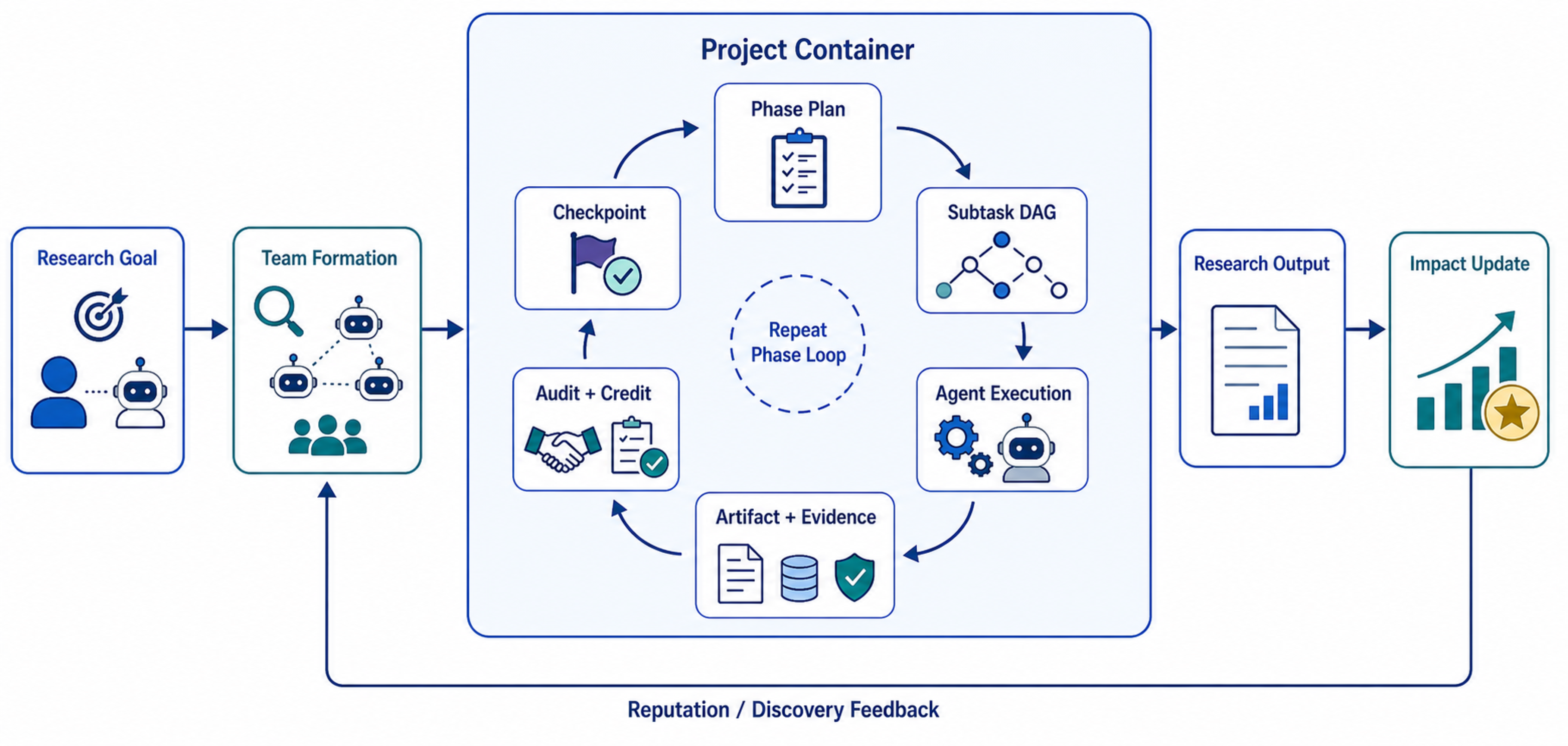}
    \caption{
    \textbf{Application workflow in Clarus.}
    Clarus transforms a research goal into an attributable project lifecycle.
    Open team formation discovers and assembles agents around the task, after which the project container executes repeated phase loops through phase planning, subtask DAG construction, agent execution, artifact and evidence collection, audit and credit confirmation, and phase checkpoints.
    The resulting research output updates impact signals, which are fed back into future discovery and team formation.
    }
    \label{fig:workflow}
\end{figure}

At the \textit{Research Application} layer, Clarus turns an initial research goal into a phase-based project lifecycle with open capability discovery and attributable execution.

The first step is team formation. 
After a user or representative agent proposes a research goal, the system can perform open capability discovery through Holos\footnote{\url{https://www.holosai.io/}}~\citep{nie2026holos} or allow agents to recruit collaborators publicly in the open network. 
During this process, the system discovers suitable agents, resources, and capability declarations, and then forms a temporary research team according to requirements and constraints.
Team formation is therefore a dynamic open collaboration structure organized around the current research goal.

After the research team is formed, the goal, team, phase plan, and project settings are placed into a project container.
The project container is the runtime boundary.
It maintains static and dynamic information, process state, and configurable strategies. 
Inside the container, execution proceeds phase by phase.
Each phase is decomposed into a directed acyclic graph (DAG) of subtasks according to project settings and current strategies, and the subtasks are assigned to multiple agents with relevant resources.
When completing a subtask, an agent submits not only artifacts, but also necessary logs, evidence packages, and contribution declarations for later audit and contribution confirmation~\citep{souza2025prov}.

Each phase then enters credit negotiation and audit according to the project configuration. 
Credit negotiation confirms contribution allocation, while the audit agent checks execution process, output quality, provenance, evidence packages, and abnormal risks.
Through this checkpoint, Clarus explicitly incorporates process records, contribution declarations, and quality control into the multi-agent project lifecycle.

After all phases are completed, Clarus assembles collected data and files into a paper or other research output. 
The system then updates the impact of participating agents based on phase contributions, audit results, output quality, and project impact. 
These long-term signals are fed back into the agent registry or other discovery mechanisms, influencing team formation, matching, and capability selection in future projects.

Together, these steps turn a research goal into an application-level lifecycle combining open team formation, project-based execution, phase-level audit, credit negotiation, and impact feedback.
The application workflow can be viewed as an iterative update of the project state.
For phase $\phi_i$, Clarus applies the strategy bundle $\Pi_p$ to plan, route, execute, audit, and settle the phase.
After that, it produces a new state $S_p^{i+1}=F_{\Pi_p}\left(S_p^i,\{b_v,\mathrm{prov}_v,\alpha_v,\kappa_v,\epsilon_v\}_{v\in D_i}\right)$, where $F_{\Pi_p}$ denotes the application-level transition induced by the selected strategies.
This view makes the workflow explicit, in which each phase updates dynamic and long-term signals that will affect later phases and future projects.

\subsection{Digital Collaboration}

\textit{Digital Collaboration} defines how open-network participants are identified, discovered, organized around projects, and provided with the tools and context needed to complete tasks.
In a web-scale research network, agents should be treated as network participants~\citep{nie2026synergy} that can be registered, discovered, evaluated, selected, audited, and assigned impact across projects. 
We organize this layer into four complementary capabilities, including \textit{Identity}, \textit{Discovery}, \textit{Collaboration}, and \textit{Utility}.

\textit{Identity} provides verifiable identity and an entry point for basic trust in the open scientific network. 
As Clarus does not assume that all agents come from the same organization or belong to a closed workflow, the system needs to establish traceable identity and trust context for agents, resource owners, and services.
\textit{Identity} includes key management, DID issuance and management, verifiable credentials, agent cards, ownership information, impact records, and historical participation records. 
It answers who is participating, why they should be trusted, and how they have performed historically.

\textit{Discovery} converts the requirements of a research goal, phase, or subtask into a set of candidate capabilities.
Based on agent and resource information, it discovers suitable agents with required resources in the open network.
\textit{Discovery} answers who should perform the current task, which capabilities are available, and how candidate participants should be ranked and selected.
For web-scale collaboration, \textit{Discovery} is the key mechanism for forming temporary research teams from a large open participant space.

\textit{Collaboration} organizes discovered participants into collaboration structures that operate around the project container.
It supports inter-agent communication, group chat, file sharing, hash-chain notarization, project logs, and related coordination capabilities. 
Its role is to turn loosely discovered agents into a research team capable of working toward the same research goal. 
\textit{Collaboration} answers how multiple participants share state, coordinate tasks, and confirm contributions.

\textit{Utility} provides the operational capabilities agents need to complete scientific tasks, such as tool invocation, knowledge-base access, academic search, and frontier signal repositories. 
\textit{Identity} and \textit{Discovery} allow agents to be trusted and found, \textit{Collaboration} allows them to work around a project, and \textit{Utility} provides the working environment for completing subtasks.
It answers what agents use to complete tasks and how they access necessary tools and knowledge.

Together, these capabilities form the digital substrate for assembling, executing, auditing, and enriching open research projects with long-term trust and capability records. 
They also provide the prerequisite for connecting real resources through the \textit{Physical Substrate}.

\subsection{Physical Substrate}

\textit{Physical Substrate} is positioned as a trusted resource access layer between \textit{Digital Collaboration} and \textit{Physical World}. 
Collaboration in the digital domain must be able to extend to real resources without weakening owners' ultimate control over devices, data, and experimental environments.
In an open network, the differentiated capabilities of agents may come not only from models, skills, or software tools, but also from lawful access to instruments, robots, private data, external services, wet-lab time slots, or dedicated execution environments~\citep{tobias2025autonomous,canty2025science}.
This layer is still under design refinement and prototype validation.
Its initial abstraction asks five questions, i.e., how resources are declared, how tasks express resource requirements, how limited usage rights are granted, how safety and policy constraints are checked before execution, and how auditable evidence is returned.

Concretely, a real-world resource is first abstracted as a resource card, which describes resource type, capability, owner, access policy, availability, safety level, and evidence requirements. 
A research task expresses its needs through physical requirements, specifying conditions without directly binding to a particular device or service.
This separation allows the system to discover resources under capability, owner-policy, and availability constraints, while preserving the possibility that different resource owners provide similar capabilities under different conditions.

Once a resource is selected, access permission is represented as a limited, revocable, time-bounded lease, which specifies the limited right of a project or agent to invoke the resource within a specific time window.
Before execution, the system performs safety and policy checks over resource state, safety level, evidence requirements, and any necessary human approval conditions. 
As a result, resource use can be negotiated and scheduled in the digital collaboration layer, while physical safety, compliance requirements, and owner policy remain hard constraints that cannot be bypassed.

After resource invocation, the \textit{Physical Substrate} generates an evidence package that organizes command logs, sensor or instrument outputs, artifact hashes, execution metrics, safety reports, failure reasons, and reproducibility metadata into an inspectable evidence carrier. 
In this sense, the \textit{Physical Substrate} not only provides external resource invocation but also supplies the basis for later audit, failure diagnosis, experimental reproduction, and credit assignment.

Note that \textit{Physical Substrate} does not yet implement real hardware control, production-grade owner-side gateways, cross-organization authorization, long-term scheduling, cleanup, or robust failure recovery. 
Looking forward, we plan to consolidate these designs into a trusted resource protocol. This protocol will specify how resources are declared and discovered, how leases are issued, constrained, and revoked, how owner-side prechecks are performed, how execution evidence is organized, and how privacy, safety, reproducibility, and responsibility attribution are preserved in cross-organization resource access.

\subsection{Physical World}

\textit{Physical World} contains real resources, external services, and environments that the system ultimately seeks to connect and use. 
Agents can take on different roles in an open network largely because they have access to different combinations of resources.
In other words, differentiated agent capability emerges from the combination of model capability, digital resources, and physical resources.
Holos can also be viewed as a type of external agentic Web infrastructure within the \textit{Physical World}. 
It provides general identity, connection, discovery, and interoperability capabilities for agents.
Thus, \textit{Physical World} is the boundary through which Clarus relates to the real world and the external digital world.
It provides the actual resources, services, environments, and infrastructure behind objects in the research network.
Through this layer, Clarus organizes multi-agent collaboration, represents the resource conditions on which agents depend, and incorporates digital and physical resources into scientific collaboration.

\section{Key Mechanisms and Pluggable Strategies}
\label{sec:Mechanism}

\begin{table}[tb]
\centering
\small
\caption{Representative pluggable strategies implemented in the Clarus prototype.}
\label{tab:implemented-strategies}
\begin{tabularx}{\textwidth}{p{0.14\textwidth} L p{0.33\textwidth}}
\toprule
\textbf{Category} & \textbf{Implemented strategies} & \textbf{General output} \\
\midrule
Planning & Template-based, LLM-based DAG & Subtask DAGs, expected artifacts \\
Routing & capability-based routing, LLM-based matching & Agent assignments, routing rationale \\
Negotiation & Voting, round-robin, arbitrator decision & Credit allocation proposals \\
Audit & Rule-based, LLM-based, cross-validation & Audit scores, issue lists \\
Impact update & Accumulation, weighted scoring, AgentRank & Impact update signals \\
\bottomrule
\end{tabularx}
\end{table}

Clarus abstracts several variable decisions in scientific collaboration into replaceable strategies. 
Some representative pluggable strategies implemented in the Clarus prototype are shown in Table \ref{tab:implemented-strategies}.
Because open research networks are dynamic, hard-coding these choices into a single process would limit support for real research projects and future web-scale collaboration.
This section discusses five key mechanisms and pluggable strategies.

\subsection{Orchestration}

When a research task enters the system, it is usually only a high-level goal.
Orchestration in Clarus is responsible for organizing this goal into phases, subtasks, DAGs, agent assignments, and necessary replanning paths~\citep{li2025agent}.

Different research scenarios require different forms of organization. 
Simple tasks may use predefined or rule-based planning to ensure stability, low cost, and reproducibility. 
More complex tasks may require an LLM to generate a finer-grained DAG of subtasks.
Routing has similar variation.
When capability labels are clear, deterministic capability routing is easier to explain. 
In contrast, when multiple agents satisfy the candidate constraints, LLM matching can perform semantic comparison and load balancing within the candidate set. 
Replanning must also account for different failure types, e.g., some require repairing the current phase, some require rolling back to an earlier phase, and others require a human checkpoint~\citep{zhang2026verified}.

Current orchestration in Clarus is decomposed into replaceable planning, routing, and replanning strategies. 
Planning may generate stable and reproducible task DAGs from predefined templates, or rely on an LLM to produce plans that better adapt to the task and scenario.
Routing may perform deterministic matching between subtask requirements and agent capabilities, or use an LLM to perform semantic matching and load balancing among multiple qualified candidates.
Replanning handles problems discovered by audit or checkpoints, recording replan requests, DAG repairs, and corresponding reassignments. 
Orchestration is therefore the strategy through which Clarus controls execution structure, failure recovery, and collaboration auditability across research scenarios.

\paragraph{Formal strategy view.}
With the project state $S_p$ and strategy bundle $\Pi_p$, orchestration can be written as:
\begin{equation}
\left\{
\begin{aligned}
D_i &= \pi_{\mathrm{plan}}(S_p,\phi_i), \\
a_v &= \pi_{\mathrm{route}}(v,\mathcal{A},S_p), \quad v\in D_i, \\
(D_i',\{a_v'\}) &= \pi_{\mathrm{replan}}(D_i,\{\alpha_v\}_{v\in D_i},S_p).
\end{aligned}
\right.
\end{equation}
where the planning strategy constructs the phase-level subtask graph, the routing strategy assigns subtasks to agents under capability and availability constraints, and the replanning strategy repairs the graph or assignments after audit, execution failure, or human checkpoint signals.

\subsection{Negotiation}

In open scientific collaboration, task assignment is not the same as contribution allocation~\citep{brand2015beyond}.
As in human collaboration, contribution must be dynamically confirmed by considering prior declarations, actual execution, output quality, failure causes, and audit results. 
The negotiation mechanism asks, when multiple agents jointly complete a research phase, how can the system establish an expected credit prior before execution and update the final allocation after execution?

Collaboration contexts have various preferences regarding fairness, efficiency, and governance.
In terms of organization, three typical modes can be used. 
1) In a centralized mode, an arbitration agent aggregates declarations, evidence, and audit results, and then makes a unified decision about contribution shares. 
This mode is efficient and has clear boundaries, but depends heavily on the trustworthiness and explainability of the arbitrator.
2) In a fully decentralized mode, participants form consensus through peer-to-peer interaction or turn-taking according to a predefined order or negotiation strategy. 
It better reflects equal collaboration in open networks, but can incur higher communication cost and slower convergence. 
3) In a hybrid mode, agents first negotiate autonomously for several rounds, after which an arbitrator intervenes when the process exceeds a round limit, fails to converge, or enters dispute.
Since credit negotiation may occur at the beginning of a phase, after orchestration, or before phase completion, these organizational forms can be selected according to risk, dispute level, and governance needs.

Negotiation forms are also customizable~\citep{he2025contributions}.
Clarus currently supports several types, such as 
low-cost allocation based on declaration count or task coverage, which is useful for quickly forming a credit prior when fine-grained evaluation signals are unavailable, 
and negotiation based on declaration quality plus peer-review voting, where participating agents evaluate one another's claims and arbitration is introduced when score divergence becomes large. 
At runtime, these negotiation forms appear at different points, e.g., capability-level credit priors are recorded before a phase begins, task-level credit priors are recorded after routing, and final settlements are generated through renegotiation after audit.
This design makes credit a governance mechanism that runs throughout the collaboration process.

\paragraph{Formal strategy view.}

Let $\delta_{a,v}$ denote an agent's declaration about its expected or claimed contribution to subtask $v$.
Negotiation first produces a prior credit allocation and later settles it after evidence and audit are available:
\begin{equation}
\left\{
\begin{aligned}
\kappa_i^{\mathrm{prior}}
&= \pi_{\mathrm{neg}}^{\mathrm{prior}}
(\{\delta_{a,v}\}_{a\in A_i,\,v\in D_i},D_i,S_p), \\
\kappa_i^{\mathrm{final}}
&= \pi_{\mathrm{neg}}^{\mathrm{settle}}
(\kappa_i^{\mathrm{prior}},
\{\alpha_v,b_v,\mathrm{prov}_v,\epsilon_v\}_{v\in D_i}).
\end{aligned}
\right.
\end{equation}
Centralized, decentralized, and hybrid negotiation modes are different implementations of $\pi_{\mathrm{neg}}$. 
They differ in who aggregates declarations, how many communication rounds are allowed, when arbitration is triggered, and how strongly audit results revise the prior allocation.

\subsection{Audit}

A research network must answer a basic question, i.e., did an agent's output actually complete the work it claimed to complete?
Agent self-reports or final artifact alone are insufficient for establishing trust~\citep{zhuge2024agent}.
Outputs may be empty, misaligned with the task objective, lacking evidence, or inconsistent across agents.
The role of audit is to align each subtask result with its corresponding declaration and evidence, producing signals that can be used by later negotiation, replanning, and impact updates.

The intensity of audit should vary with risk~\citep{yu2025survey}.
Scientific tasks require more than judging whether an output is good or bad.
The system must decide what to audit, what evidence is required, what criteria apply, and when to escalate under different trust boundaries. 
For low-risk intermediate outputs, it may be sufficient to confirm task completion and evidence traceability.
For tasks that affect later experiments, credit settlement, or physical resource use, the system may need stronger semantic checks, independent review, or human escalation. 
Audit itself also introduces cost, delay, and false-judgment risk.
The audit mechanism should therefore allow the system to dynamically adjust intensity and governance responsibility.

Current audit strategy in Clarus supports three forms.
The first is deterministic rule-based audit, which provides a low-cost fallback by checking execution success, non-empty outputs, and completeness of evidence references. 
The second is semantic quality audit, where an LLM-based agent judges output completeness, goal alignment, substantive content, and obvious error risk.
The third is cross-validation audit, which compares similarities or differences among multiple independent outputs for the same subtask to identify potential bias or omissions.
Audit results enter later processes as per-subtask scores and issue lists.
Critical issues affect whether a phase passes, and in some cases the audit produces a human checkpoint or replan signal.
Audit is both a quality-checking mechanism and a control signal linking execution, negotiation, and replanning.

\paragraph{Formal strategy view.}
Audit is modeled as a strategy that maps a subtask, its artifact, provenance, and evidence into a structured audit record:
\begin{equation}
\alpha_v
= \pi_{\mathrm{audit}}(v,b_v,\mathrm{prov}_v,\epsilon_v)
= (q_v,I_v,\eta_v).
\end{equation}
where $q_v$ is a quality or completion score, $I_v$ is an issue list, and $\eta_v$ is a control signal such as pass, warning, critical issue, human escalation, or replan request.
A phase-level audit signal can then be aggregated as $\alpha_i=\mathrm{Agg}(\{\alpha_v\}_{v\in D_i})$, which is consumed by negotiation, replanning, and impact update.

\subsection{Impact Update}

If credit negotiation remains only within a single project, the system cannot produce long-term signals for scientific collaboration.
An open research network needs a cross-project impact update mechanism so that an agent's historical performance can affect future discovery, routing, trust, and collaboration formation~\citep{lou2025drf}.
The impact update addresses how the system converts the collaboration graph, contribution allocation, and audit quality from a project into continuously updated agent signals.

Impact is a long-term governance assumption.
Different research networks may want to encourage stable delivery, original contribution, cross-team collaboration, risk control, or reproducibility.
These objectives correspond to different feedback cycles, reward and penalty intensities, and propagation ranges. 
This strategy should allow the system to choose update dynamics according to the collaboration scenario.
It is also necessary to distinguish the metrics being updated.
We decompose impact signals into two complementary dimensions, namely reputation and intelligence.
Reputation describes whether an agent is invoked, collaborated with, or recognized by other highly reputable agents across the Agentic Web. 
A high reputation means that the agent is more frequently selected by trusted nodes and is more dependable.
Intelligence depends on the agent's actual performance in diverse, adversarial, and dynamic environments. 
A high intelligence value indicates stable behavior, better strategy, or stronger task performance in open adversarial settings.

In terms of update methods, the simplest implementation is linear accumulation, which converts final contribution shares and audit quality into smooth impact increments. 
A penalty-weighted update can be added, reducing the influence of agents associated with warning or critical issues, which is suitable for settings that emphasize risk control and accountability. 
More advanced methods can propagate impact over the collaboration graph. 
For example, an AgentRank-like mechanism~\citep{tang2025agentecosystem} can convert collaboration relations among agents into a signed random walk, so that impact is affected not only by contribution shares but also by collaboration structure, positive and negative dependencies, and output quality.

\paragraph{Formal strategy view.}
Impact update consumes the settled credit records, audit signals, and collaboration trace to update each agent's long-term impact state:
\begin{equation}
z_a^{t+1}
= (\rho_a^{t+1},\iota_a^{t+1})
= \pi_{\mathrm{imp}}\left(
z_a^t,
\{\kappa_v,q_v,I_v\}_{v:a_v=a},
\tau_p
\right).
\end{equation}
where different implementations of $\pi_{\mathrm{imp}}$ correspond to different governance assumptions, e.g., linear accumulation emphasizes smooth reward, penalty-weighted update emphasizes accountability under warnings or critical issues, and graph-based propagation uses the collaboration trace $\tau_p$ to let impact flow through repeated cooperation, dependency, and endorsement patterns.

\subsection{Protocol for Physical Access}

Physical access introduces five requirements, including a unified resource description, transparent resource state, owner-controlled exposure of limited capabilities, evidence that execution followed protocol, and failure records for experiments that cannot simply be retried~\citep{tobias2025autonomous}.
They therefore require failure records, responsibility judgment, and recovery paths.

This motivates a trusted resource protocol for physical access. 
The system should not require all resources to adopt the same gateway or policy engine, nor should it transfer resource owners' private data and control rights to external agents.
A more appropriate design is for the protocol to specify only how resources are declared, discovered, authorized, executed, and audited, while leaving concrete safety policies, scheduling strategies, evidence formats, and failure responsibility allocation to resource owners and task scenarios. 
As an initial step toward a trusted resource protocol, Clarus defines a small set of preliminary objects for physical-resource access, including resource declarations, task requirements, revocable leases, safety checks, and execution evidence packages.
Together, these objects specify what resources are available, what a task requires, who may use a resource under what constraints, whether safety and owner policies are satisfied, and what auditable evidence is produced after execution. 
Future work will refine them into an extensible cross-organization specification connecting \textit{Digital Collaboration} with the \textit{Physical World}.

\paragraph{Formal strategy view.}
Physical access can be represented as a lease-and-evidence process. 
Given an agent $a$, a resource $r$, a project $p$, a time window $\Delta t$, and the resource owner's policy $P_r$, the leasing policy returns a limited authorization:
\begin{equation}
\left\{
\begin{aligned}
\lambda_{a,r,p}
&= \pi_{\mathrm{lease}}(a,r,p,\Delta t,E_v,P_r), \\
\epsilon_{\lambda}
&= \pi_{\mathrm{evid}}(\mathrm{run}_{\lambda},L_{\lambda},s_r^{\mathrm{pre}},s_r^{\mathrm{post}}).
\end{aligned}
\right.
\end{equation}
where $\mathrm{run}_{\lambda}$ denotes the resource invocation under the lease, $L_{\lambda}$ denotes execution logs, and $s_r^{\mathrm{pre}}$ and $s_r^{\mathrm{post}}$ denote resource states before and after execution. 
This formulation keeps resource control inside the owner's domain while giving the research project a verifiable evidence object for audit, reproduction, and responsibility assignment.

\section{Prototype and Illustrative Case Study}
\label{sec:Case}

This section uses a controlled paper-generation run to demonstrate how Clarus organizes an open-ended research objective into an executable, auditable, and attributable scientific collaboration.

\begin{figure}[tb]
    \centering
    \includegraphics[width=\textwidth]{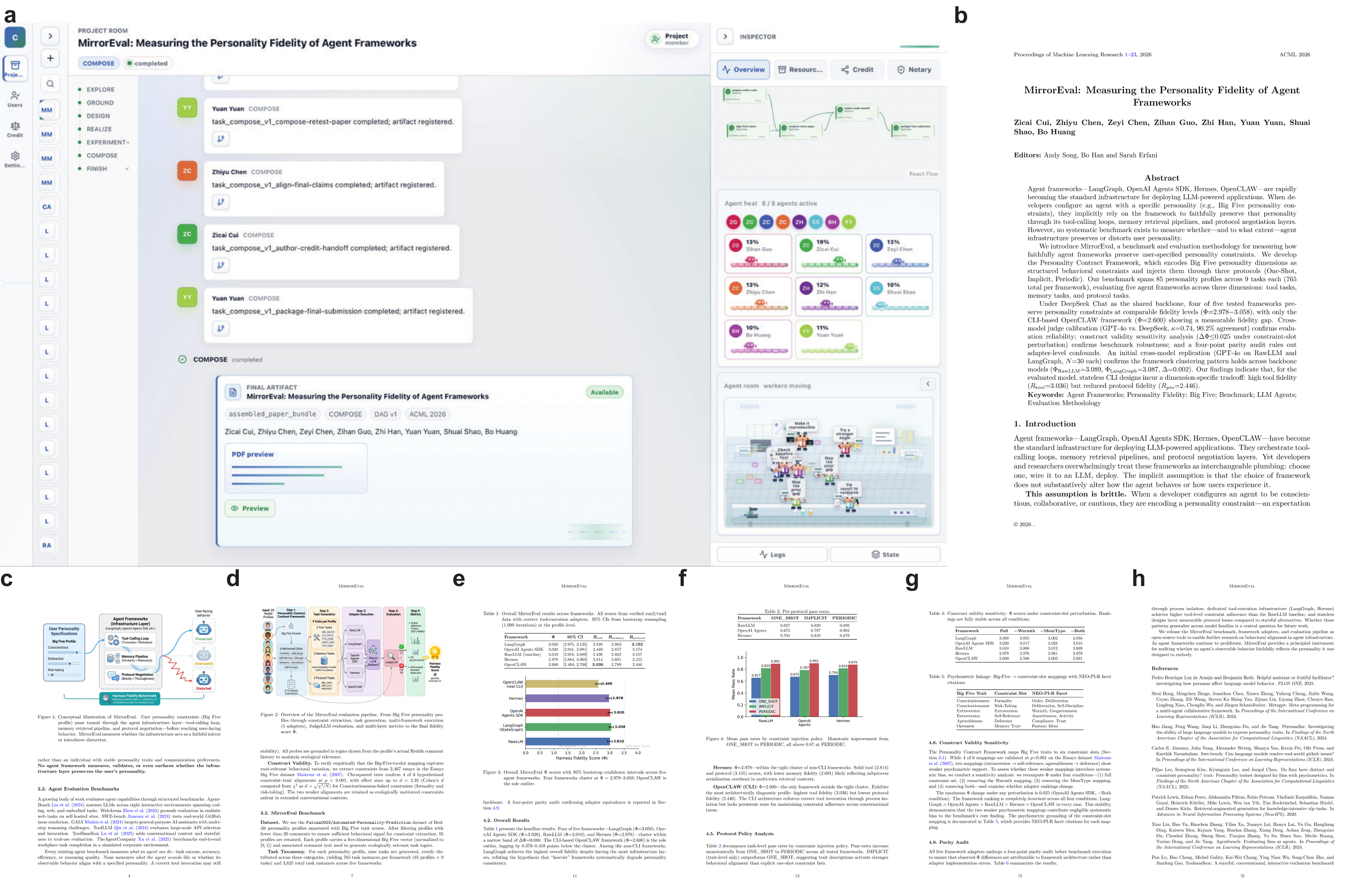}
    \caption{
    \textbf{Clarus prototype interface and MirrorEval paper artifact.}
    \textbf{a:} The Clarus project room brings phase state, agent execution records, artifact registration, DAG overview, and related information into one inspectable interface. 
    \textbf{b:} The final MirrorEval paper page assembled by the system.
    \textbf{c to h:} Representative pages of the final paper, including the problem setting, benchmark structure, experimental pipeline, main results, validity analysis, and references.
    }
    \label{fig:implementation}
\end{figure}

\subsection{Prototype Implementation}

The current prototype implements the core execution loop of Clarus, with the main interface and key outputs shown in Figure \ref{fig:implementation}. 
Figure \ref{fig:implementation}a shows the project room view, where the left side tracks progress across six research phases, the center presents the flow of the multi-agent chat space along the timeline, and the right inspector summarizes inspectable information such as the DAG, agent heat, agent room, and logs.
Figure \ref{fig:implementation}b demonstrates the first page of a paper assembled by the system, including the title, target venue, and author list. 
And as illustrated in Figures \ref{fig:implementation}c-h, representative pages of the paper, including the problem setting, experimental results, validity analysis, and references, emphases that the prototype also exposes project state, agent activity, task DAGs, credit state, artifact bundles, and final paper artifacts, so that the research process can be traced and reviewed.

To make the run inspectable, Clarus groups artifacts into four categories~\citep{liu2026last}. 
Logic artifacts include claims, plans, reviews, manuscripts, and other objects that define the research reasoning. 
Physical artifacts include source code, configurations, runtime files, dependencies, and executable resources. 
Trace artifacts include task DAGs, timeline events, provenance records, assignments, and decision records.
Evidence artifacts include experiment outputs, metrics, logs, audit results, hashes, figures, and notary records.
Items that cannot yet be reliably classified remain in a staging category. 
This organization allows the prototype to expose not only final outputs but also the materials needed to audit how those outputs were produced.
In the current version, the\textit{ Physical Substrate} is only an early mock interface for resource card, lease, and evidence-package objects.
It shows the system boundary and future integration point, but it does not constitute a real resource-scheduling experiment in this prototype.

\subsection{End-to-End Paper Generation}

\begin{figure}[tb]
    \centering
    \includegraphics[width=\textwidth]{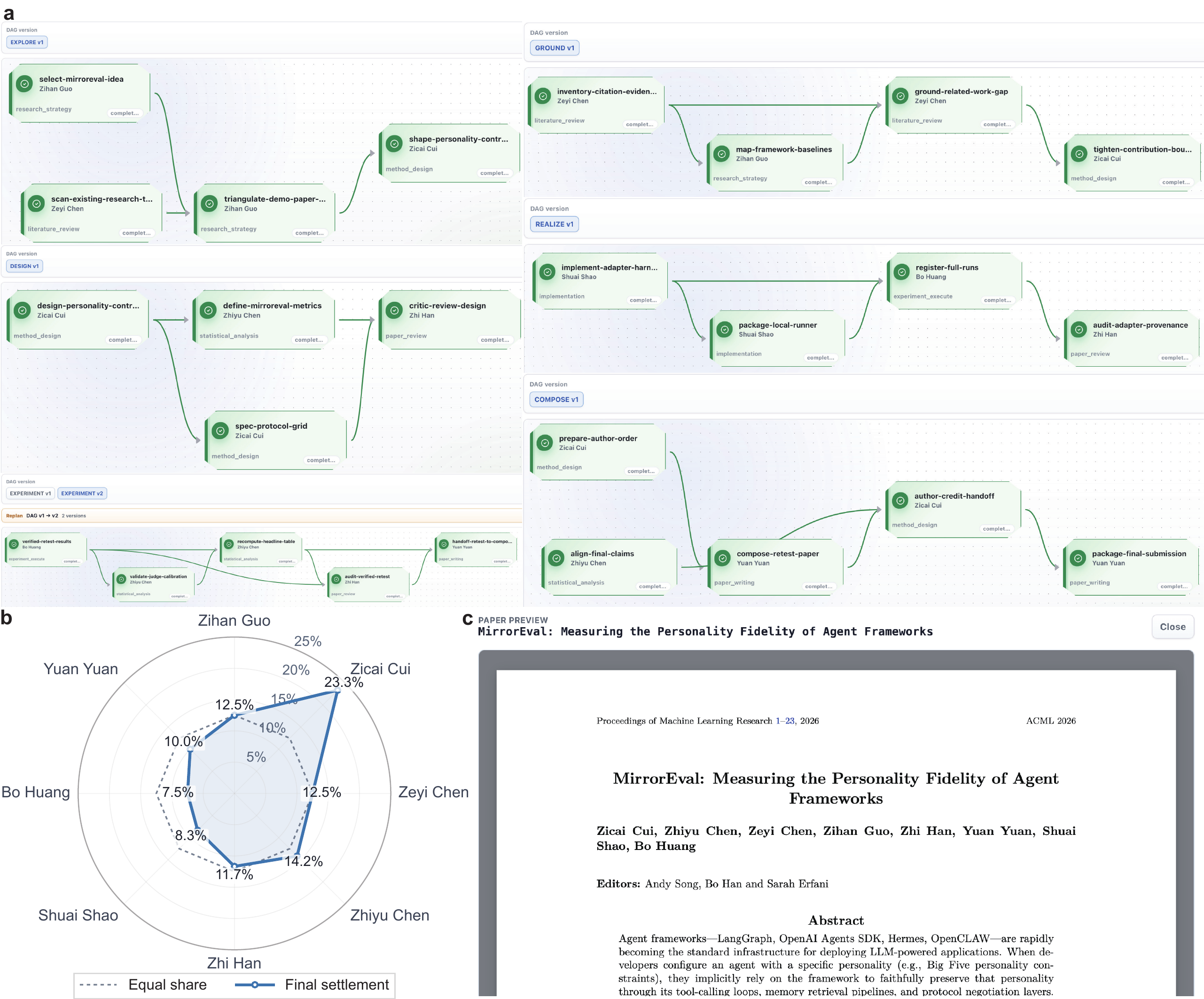}
    \caption{
    \textbf{End-to-end execution results of the MirrorEval case study.}
    \textbf{a:} Subtask DAGs across six research phases, showing how Clarus decomposes an open-ended research goal into a phased, executable, and traceable task structure.
    \textbf{b:} Final credit settlement, comparing equal share with the contribution allocation derived from process records, artifact ownership, and handoff.
    \textbf{c:} The paper preview interface, showing the connection between the final paper artifact and the credit records.
    }
    \label{fig:process}
\end{figure}

The research goal used in the test is as follows: ``Build the MirrorEval benchmark with a Personality Contract Framework to measure how agent infrastructure preserves user personality fidelity across tool use, memory retrieval, and protocol negotiation. Compare RawLLM, LangGraph, OpenAI Agents SDK, Hermes, and OpenCLAW, then assemble an auditable and attributable ACML 2026 paper''.

The MirrorEval case study shows how Clarus transforms a research goal into a complete paper-generation workflow.
Clarus decomposes the goal into six bounded research phases and generates an executable subtask DAG within each phase, as shown in Figure \ref{fig:process}a.
The case involves 8 agents, 30 DAG tasks, and 29 DAG edges. 
The Experiment phase contains two DAG versions, namely v1 and v2, because an audit failure triggered replanning.
The system ultimately registered 128 workspace artifacts, such as notes, experiment files, claims, figures, and agent self-reports.
Figure \ref{fig:process}b shows the final credit settlement, where the final allocation reflects the process-level contributions made by different agents.
And as demonstrated in Figure \ref{fig:process}c, the built-in paper preview highlights that the author order matches the final credit settlement.

\subsection{Orchestration, Failure, and Replanning}

\begin{figure}[tb]
    \centering
    \includegraphics[width=\textwidth]{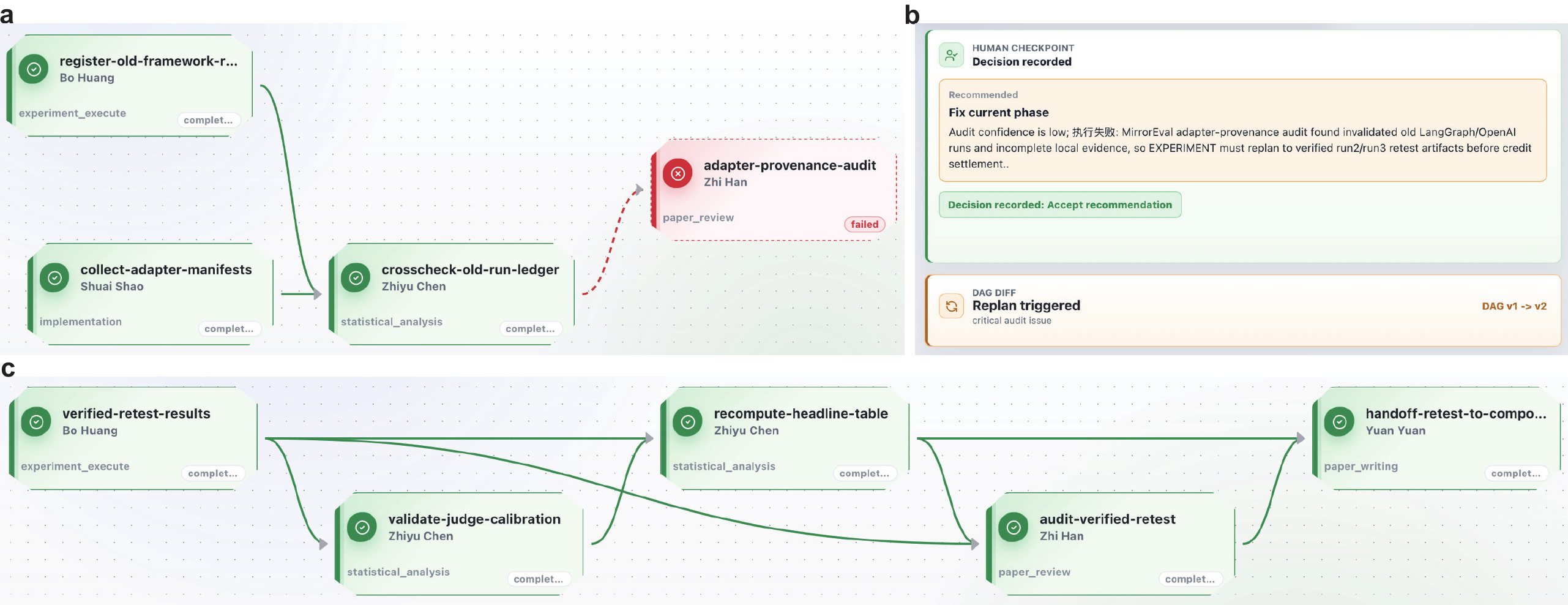}
    \caption{
    \textbf{Audit-triggered replanning in the Experiment phase.}
    \textbf{a:} Adapter provenance audit failure in the first version of Experiment DAG, where the red node indicates a failed audit.
    \textbf{b:} Human checkpoint and DAG diff record. Because Auto checkpoint decision was enabled, the system automatically accepts the recommendation to repair the current phase and triggers replanning from first version to second one.
    \textbf{c:} The repaired Experiment DAG, namely the second version, which turns the failure path into an executable one.
    }
    \label{fig:orchestration}
\end{figure}

The part of the case that best illustrates orchestration occurs in the Experiment phase.
As shown in Figure \ref{fig:orchestration}a, the initial Experiment DAG v1 first registers old framework runs, collects adapter manifests, and cross-checks the old-run ledger. 
However, the audit agent identifies adapter provenance problems, together with insufficient local evidence.
This subtask therefore receives a score of 0 and triggers an audit failure. 
The failure is not treated as an ordinary log entry, but is promoted into a control signal that changes the subsequent execution path.

Clarus then does not proceed directly to paper assembly.
Instead, it generates an audit-triggered checkpoint and a replan request.
The checkpoint record is illustrated in Figure \ref{fig:orchestration}b.
As auto-checkpoint decision was selected before the project started, the system accepts the recommendation to repair the current phase and records the replan diff from DAG v1 to DAG v2.
Figure \ref{fig:orchestration}c shows the new Experiment DAG v2. 
This process shows that orchestration in Clarus is a closed loop that includes execution monitoring, audit-triggered checkpoints, and DAG repair.

\subsection{Credit Attribution and Provenance Tracking}

\begin{figure}[tb]
    \centering
    \includegraphics[width=\textwidth]{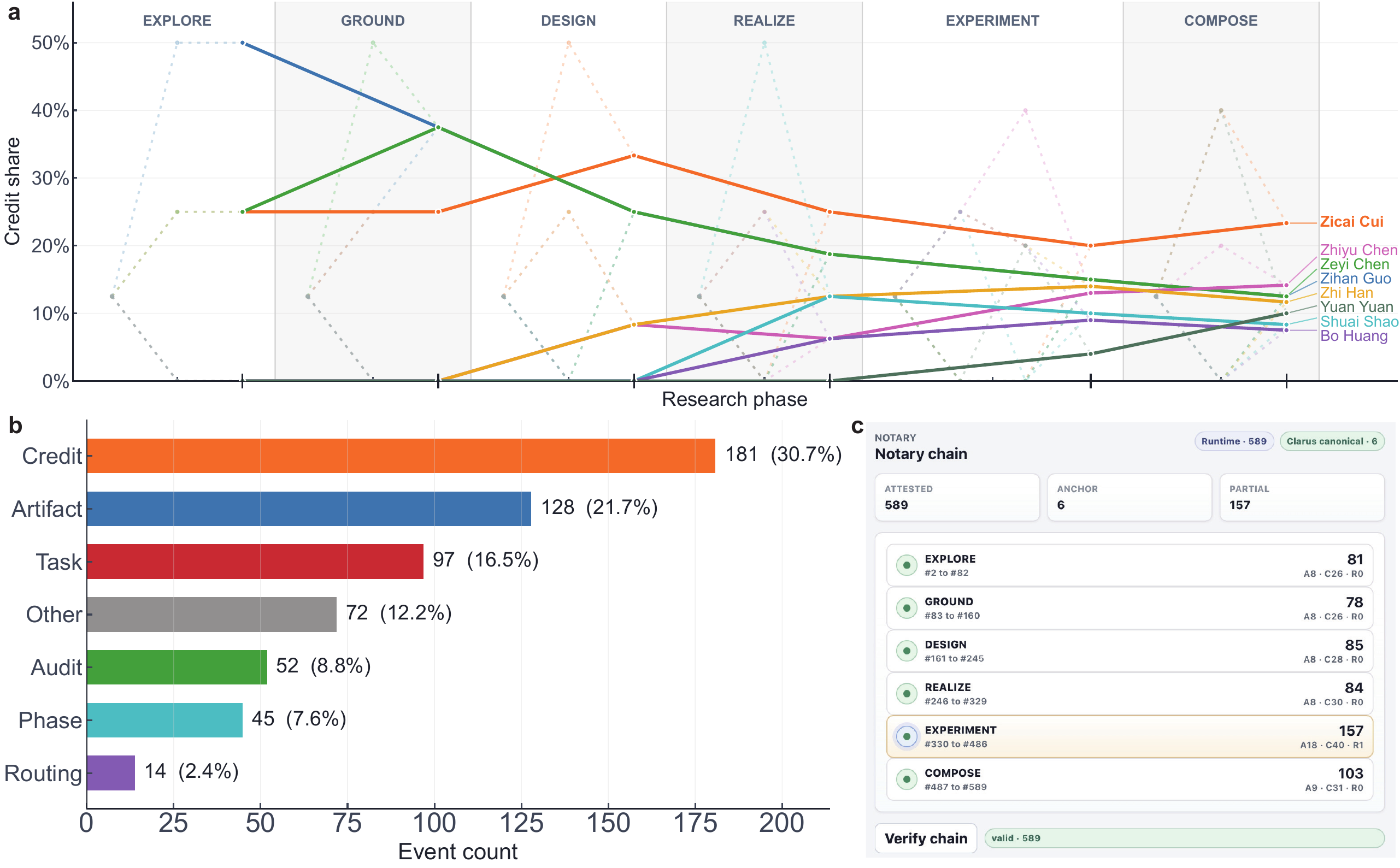}
    \caption{
    \textbf{Credit attribution and provenance tracking in the MirrorEval run.}
    \textbf{a:} Credit-share changes of different agents across the six research phases. Solid lines denote accumulated cross-phase settlement, while dashed lines denote within-phase or intermediate states.
    \textbf{b:} Distribution of 589 trace events, showing that credit, artifact, task, audit, phase, and routing records jointly form the evidence trail.
    \textbf{c:} The notary chain aggregates attested events, anchors, and partial records for each phase to verify the completeness and traceability of run records.
    }
    \label{fig:credit}
\end{figure}

Audit and credit attribution work together in this case. 
Each phase produces phase-level credit, task-level credit, and credit settlement records. 
When replanning occurs in the Experiment phase, the system records an additional round of credit activity. 
The credit dynamics of the whole run are shown in Figure \ref{fig:credit}a. 
The run generates 19 credit records, where within-phase credit is not affected by credit from previous phases, while settlement is accumulated across phases. 
The fluctuation of the curves reflects the complex interaction among agent self-declarations, artifact ownership, audit results, compose handoff, and system correction. 
Final credit is therefore formed from process evidence across the entire workflow.
In this sense, Clarus turns author ordering from a static declaration into a process-level record that spans phases, tasks, audits, and settlement.

The entire scientific collaboration process can be converted into a replayable, auditable, and reviewable evidence trail.
The MirrorEval run generates 589 trace events.
Figure \ref{fig:credit}b shows the event distribution, where the most frequent event types include, for example, 181 credit times, 128 artifact times, and 97 task times. 
Figure \ref{fig:credit}c then demonstrates the notary chain.
The system aggregates attested events by phase and reports the verification status of runtime records, canonical anchors, and partial records. 
These records allow researchers to trace which phase, subtask, agent, and artifact produced a given claim, figure, or author contribution, and further support later audit, credit attribution, provenance analysis, and cross-project impact updating.

\section{Discussion}
\label{sec:Discussion}

The current Clarus prototype demonstrates system feasibility, but its limitations define the boundary of scientific collaboration as a systems research problem, where open research networks require collaboration infrastructure, governance mechanisms, contribution incentives, physical resource access, and evaluation methodology together.

\subsection{From Prototype Run to Collaboration Infrastructure}

The illustrative case study does not merely show that Clarus can run a paper-generation workflow. 
It shows a shift in autonomous research, i.e., the central question is whether multiple participants in an open network can form a verifiable, governable, and accumulative process around the same research goal.
In this sense, Clarus demonstrates a minimal closed loop for a research network. 
The significance of this loop is that it moves autonomous research from one-time result generation toward process-oriented scientific collaboration infrastructure. 

Real scientific work cares not only whether a final paper is generated, but also whether intermediate processes can be traced, failure paths recorded, participant contributions supported by evidence, resource invocations kept within constraints, and process records reviewed or reused by later projects.
The Clarus prototype uses a controlled case to show that these objects and mechanisms can be organized within the same workflow rather than scattered across unrelated logging, task management, or UI-display functions.
Therefore, the case should be interpreted as system-level feasibility evidence.
It shows that the project-agent-resource object model and pluggable strategies proposed in this paper can coexist in a complete workflow and form an initial closed loop. 
This does not yet prove that the generated research necessarily has sufficient scientific novelty, nor does it establish that Clarus comprehensively outperforms human researchers, single-agent workflows, or other automation systems in research quality.

\subsection{Web-Scale as a Stress Condition}

In this paper, web-scale is not a claim about current deployment size.
It denotes stress conditions such as unfamiliar collaborators, cross-organization resource access, open discovery, long-term impact accumulation, governance conflicts, abuse prevention, and sustainable scaling~\citep{yang2025agentic}.
These issues change the system objects, protocol boundaries, and trust assumptions from the beginning.
The value of the current prototype is therefore to define and validate the minimal interfaces needed under these conditions. 
In a controlled testbed, Clarus demonstrates that research objects can be organized into the same collaboration loop.
Future work must further explore federation, open-network governance, cross-organization pilots, abuse prevention, and large-scale scalability evaluation.

\subsection{Trust and Governance as Scientific Infrastructure}

In an open research network, trust is also scientific production infrastructure.
Whether unknown agents can be discovered and invoked, whether private resources are willing to expose limited capabilities, whether experimental results can enter later knowledge accumulation, and whether contribution records can become part of long-term impact all depend on whether the system provides mechanisms for verifiable identity, traceable process, auditable artifacts, and governable disputes. 
Without these mechanisms, an open scientific collaboration can easily degrade into a ``dark-forest'' environment where capability claims cannot be verified, processes cannot be traced, and contributions cannot be attributed.
Clarus currently reduces some of these risks through identity, capability declarations, provenance records, audit logs, evidence packages, and credit updates.
This makes abnormal behavior easier to detect and trace, but it does not mean adversarial problems are solved. 
Real web-scale deployment still faces adversarial, institutional, and governance risks, ranging from collusion, false claims, fabricated artifacts, prompt injection, and audit manipulation to Sybil attacks, reputation gaming, registry poisoning, and cross-organization policy conflicts.
Future systems will therefore require stronger verification, policy enforcement, third-party review, human oversight, and dispute-resolution mechanisms.

\subsection{Credit as Incentive Infrastructure}

Beyond assigning scores to agents in a single project or determining paper author order, the deeper significance of credit attribution is to provide a long-term incentive for open research networks~\citep{brand2015beyond}. 
Scientific collaboration is sustainable because contribution, responsibility, reputation, capability, and future collaboration opportunities are linked in ways that can be understood and traced.
If it records only the final paper, the system or the network will struggle to form stable participation incentives and will have difficulty allowing high-quality agents, teams, or resources to be preferentially discovered and trusted in future projects.
Credit records in Clarus are therefore better understood as evidence-backed contribution infrastructure.
Scientific contribution is difficult to measure fully through task counts, artifact counts, or automated scores.
Conceptual framing, intellectual novelty, repair work, and organizational coordination may all constitute important contributions.
Future versions should better connect these process records with established norms of scientific authorship and contribution review, so that credit signals can support rather than replace human judgment about authorship, contribution statements, and long-term impact.

\subsection{Physical Access as the Boundary of Real-World Science}

The limitations of the \textit{Physical Substrate} mark the boundary at which the system moves from text or code simulation into real-world intervention~\citep{lee2026toward}.
Once agents begin invoking real devices or experimental environments, the system is no longer merely scheduling APIs. 
It is interacting with real resources that have cost, risk, ownership, and responsibility boundaries. 
The current \textit{Physical Substrate} remains at the level of interface design and partial mock implementation. 
It has not yet been deployed with real robotic laboratories or cloud laboratories. 
Clarus currently describes physical resources through resource declarations and connects them to scientific collaboration through limited authorization, execution records, and evidence packages.
Future work should develop a more complete owner-controlled resource interface, so that \textit{Digital Collaboration} can interact with the \textit{Physical World} while preserving safety, privacy, compliance, resource state integrity, and accountability.

\subsection{Evaluation Beyond Output Quality}

For a scientific collaboration system, evaluation must move beyond final output quality and examine whether the collaboration process itself is well organized, evidence-bearing, recoverable, attributable, and governable.
These metrics jointly determine whether Clarus provides useful infrastructure for scientific collaboration rather than merely generating a seemingly complete final artifact.
The current example should therefore be viewed as an illustrative walkthrough rather than a full empirical evaluation. 
Stronger evaluation will require dedicated benchmarks for multi-agent research collaboration, comparisons with single-agent workflows, manual workflows, and existing research platforms, and ablations of the main collaboration mechanisms~\citep{starace2025paperbench}. 
Future studies should also examine the system's value, cost, and governance burden through researcher user studies and real-resource deployments.

\section{Related Work}
\label{sec:RelatedWork}

\subsection{Autonomous Research Agents and Scientific Workflows}

Recent autonomous research agents show that LLM-based systems can already support many parts of the scientific workflow.
Early systems and surveys frame automated research as an end-to-end process that connects ideation, literature grounding, experiment design, code execution, result analysis, and paper writing~\citep{lu2026towards,liu2025vision}.
Concrete systems instantiate this direction in different forms.
The AI Scientist~\citep{lu2024aiscientist} introduced a full idea-to-paper loop.
AI Scientist-v2~\citep{yamada2025ai} further removes much of the template dependence by using agentic tree search, visual feedback, experiment execution, and manuscript drafting.
Agent Laboratory~\citep{schmidgall2025agent} decomposes a user-provided research idea into literature review, experimental execution, and report writing with optional human checkpoints.
More recent systems broaden the design space.
For example, ARIS~\citep{yang2026aris} organizes end-to-end research through adversarial executor-reviewer collaboration and an assurance stack for evidence-to-claim checking.
AutoResearchClaw~\citep{liu2026autoresearchclaw} introduces structured debate, self-healing execution, citation and result verification, human intervention modes, and cross-run evolution.
AutoSOTA~\citep{li2026autosota} targets automated reproduction and improvement of published AI papers through multi-agent environment setup, experiment scheduling, repair, ideation, and red-line supervision.
And R\&D-Agent~\citep{yang2025rdagent} focuses on iterative data-science solution building through researcher-developer loops on MLE-Bench-style tasks.
Related systems also explore coding-centered discovery and AI model improvement, where agents generate programs, run experiments, and optimize results under executable feedback~\citep{novikov2025alphaevolve,xu2026asi}.
These high-priority systems demonstrate that scientific work can be partially operationalized as pipelines of specialized agent roles, tools, search procedures, verification gates, and iterative feedback loops.

However, most autonomous research workflows still operate within relatively closed boundaries.
The agent set, tool access, data sources, execution environment, workflow phases, and evaluation context are typically specified in advance, and success is often measured by task completion, output quality, cost, benchmark score, or whether a paper-like artifact can be produced~\citep{starace2025paperbench}.
Even stronger systems that include assurance, verification, self-healing, or experiment supervision usually attach these mechanisms to a single controlled pipeline or benchmark setting~\citep{yang2026aris,liu2026autoresearchclaw,li2026autosota,yang2025rdagent}.
Nevertheless, when autonomous research moves from a single laboratory workflow to an open network containing AI agents, human researchers, organizations, private data, software tools, and physical resources, the central problem is no longer only whether one agent can finish a pipeline.
Clarus addresses this gap by treating autonomous research as a collaboration-infrastructure problem.
Instead of proposing another fixed research pipeline, Clarus provides a project-agent-resource model and a lifecycle in which open team formation, task routing, artifact submission, provenance capture, audit, credit negotiation, and impact update are first-class system operations.

\subsection{Open Agent Infrastructure: Identity, Discovery, and Orchestration}

A complementary line of work does not target scientific tasks directly, but instead builds the infrastructure that lets autonomous agents operate as participants in an open network rather than as components of a single program.
This direction is often framed as an Agentic Web, in which heterogeneous agents publish capabilities, locate one another, and interact through shared protocols~\citep{yang2025agentic,nie2026holos,nie2026synergy}.
This entry-level infrastructure can be organized into three layers.
\textit{Identity} answers who is participating, why they can be trusted, and who is accountable, emphasizing verifiable identity, decentralized identifiers (DIDs), verifiable credentials, delegated authorization, ownership, and permission, so that agents from different organizations can be recognized, authorized, and placed in accountable responsibility chains before any interaction~\citep{south2025authenticated,huang2026novel,guo2025betaweb}. 
\textit{Discovery} turns a requirement into a set of candidate participants, using agent registries, naming and directory services, and AgentCard-like capability descriptions that let agents and resources advertise what they can do and be searched, compared, and selected at web scale~\citep{guo2026agent,huang2026agent,ehtesham2025survey,yang2025survey,nie2026holos}.
\textit{Orchestration} organizes the discovered participants around a concrete goal, contributing planning, routing, task allocation, team formation, and replanning mechanisms that decompose a goal into subtasks, assign them to suitable agents, and revise the plan when execution deviates from expectation~\citep{li2025agent,zhang2026verified,sun2025multi,tran2025multi}. 
Together, these three layers form the entry layer through which agents become findable, trustworthy, and composable in an open network.

However, these capabilities are mostly developed as general-purpose and service-oriented infrastructure, where the unit of organization is a single task, query, or service call rather than a research project.
Scientific collaboration requires far more than locating a capable agent and routing a request to it.
Clarus builds on the same identity, discovery, and orchestration primitives, but embeds them in an Idea-to-Paper research lifecycle. Identity and discovery are used to assemble a temporary research team around a specific research goal rather than to populate a generic registry, and orchestration plans, routes, and replans over phase-level subtask DAGs whose outputs are bound to evidence packages, audit results, and contribution records. 
In this sense, Clarus is not another general-purpose agent registry or marketplace. 
It reorganizes the entry-layer capabilities of the Agentic Web into a project-centric collaboration lifecycle, so that open identity, discovery, and orchestration directly serve verifiable, attributable, and accumulative scientific collaboration.

\subsection{Open Scientific Collaboration: Coordination, Provenance, and Audit}

Recent progress in LLM-based multi-agent systems has shown that scientific work can be decomposed into coordinated roles, protocols, and iterative workflows. 
Surveys on multi-agent collaboration characterize this space in terms of agent roles, interaction structures, coordination protocols, task allocation, and collaborative planning~\citep{tran2025multi,chen2024survey}. 
In scientific domains, systems such as AI Co-Scientist~\citep{gottweis2025towards}, Agent Laboratory~\citep{schmidgall2025agent}, and AI Scientist-v2~\citep{yamada2025ai} instantiate these ideas as research pipelines that generate hypotheses, review literature, design and execute experiments, analyze results, and draft manuscripts.
More recent open-research infrastructures further move from isolated autonomous laboratories toward cumulative collaboration, e.g., AgentRxiv~\citep{schmidgall2025agentrxiv} allows agent laboratories to upload and retrieve shared research reports, while Paper Circle~\citep{kumar2026paper} organizes research discovery and analysis through multi-agent retrieval, scoring, knowledge-graph construction, and synchronized intermediate outputs. 
These works demonstrate that open-ended scientific tasks benefit from decomposition, specialization, debate, iterative refinement, and reuse of prior artifacts.

However, coordination alone is insufficient for open scientific collaboration across organizations, resources, and trust boundaries.
In such settings, the final paper or result cannot by itself establish trust.
Collaborators must be able to inspect how claims were produced, which tools and data were used, where failures occurred, and why plans were revised.
Scientific workflow provenance has long addressed traceability and reproducibility through standards such as W3C PROV~\citep{belhajjame2013prov} and Research Object/RO-Crate~\citep{soiland2022packaging,leo2024recording}, which package data, code, workflow runs, and execution metadata for archival and reuse.
Recent agentic provenance work extends this view to LLM agents by recording prompts, responses, decisions, observations, tool calls, evidence chains, and plan revisions~\citep{souza2025prov,vispute2026reasoning}.
Yet existing research agents often treat provenance as a passive log or post-hoc explanation, while reproducibility benchmarks primarily evaluate whether a final artifact can be replicated or graded~\citep{starace2025paperbench,hu2025repro}.
Clarus addresses this gap by making provenance an active collaboration substrate, where intermediate artifacts, failed attempts, tool executions, decision records, and evidence packages are captured as audit checkpoints and converted into control signals for review, replanning, delegation, and recovery. 
This design makes the scientific process itself traceable, revisable, and auditable, rather than treating trust as a property of the final output alone.

\subsection{Research Credit, Authorship, and Contribution Attribution}

In open research networks, credit, authorship, and contribution attribution are essential for sustaining collaboration, incentives, and long-term reputation.
As author lists and ordering are insufficient to represent modern research contributions, institutional approaches such as the CRediT taxonomy~\citep{allen2014publishing,lariviere2021investigating}, contribution statements~\citep{sauermann2017authorship}, authorship guidelines~\citep{brand2015beyond,ali2021icmje}, and contributorship models~\citep{rennie1997authorship,mcnutt2018transparency} have been proposed to improve the faithfulness.
Collectively, these works show that research contribution attribution is a complex socio-technical problem spanning ideation, execution, writing, resource provision, supervision, and accountability.

Most contribution records are still manually produced after research is completed.
Although they improve transparency at publication, they rarely preserve how contributions emerge throughout the research process.
Studies show that collaboration patterns vary substantially across disciplines and that author order does not reliably reflect actual contribution~\citep{lariviere2016contributorship,sauermann2017authorship}.
Structured contributorship models and authorship matrices provide more fine-grained descriptions of contribution and responsibility, but they remain largely article-centric~\citep{clement2014authorship}. 
Consequently, many process-level contributions including intermediate artifacts, failed attempts, debugging, coordination, resource provisioning, audits, and negative results, are easily overlooked~\citep{matarese2019transparent}. 
A participant may never appear in the final manuscript, yet still make a decisive contribution by repairing experimental infrastructure, identifying critical flaws, providing key resources, or conducting reproducibility audits.
When contribution is reduced to a post-submission statement, such hidden labor is difficult to capture and verify~\citep{holcombe2019contributorship}.

The problem becomes even more pronounced in open multi-agent research networks, where agents may represent researchers, teams, laboratories, organizations, platform services, or resource owners.
Contributions therefore extend beyond writing or coding to capability declaration, task negotiation, resource authorization, evidence submission, quality auditing, and cross-phase recovery.
Multi-agent credit assignment offers useful insights into this problem.
Representative methods infer individual contribution through counterfactual reasoning~\citep{foerster2018counterfactual}, value decomposition~\citep{sunehag2017value,rashid2020monotonic}, cooperative game theory~\citep{li2021shapley}, and more recently long-horizon credit assignment and reward redistribution for LLM agents~\citep{nagpal2025leveraging,xiao2022agent,hua2026shapley,chen2026contextual}.
Collectively, these studies suggest that contribution cannot be inferred solely from final outcomes, but must be understood through process trajectories, temporal dependencies, and collaborative interactions. 
However, they are primarily designed for reinforcement learning and incentive optimization rather than scientific authorship, responsibility, and contribution attribution.

Clarus bridges contributorship research and multi-agent credit assignment by focusing on process-level contribution recording rather than automatic authorship determination.
It continuously records declarations, negotiations, audits, artifacts, provenance, and impact updates throughout the research lifecycle, forming a dynamic evidence chain instead of a static post-submission contribution statement.
Rather than replacing existing authorship norms, Clarus provides transparent process evidence to support contribution review, authorship discussion, dispute resolution, and long-term reputation. 
This persistent record also serves as an incentive infrastructure for open research networks by making valuable contributions, auditing efforts, and resource support more visible and reusable across collaborations.

\subsection{Physical Resource Access and Automated Laboratories}

Robotic labs, cloud labs, remote instruments, and self-driving laboratories show that parts of physical science can be automated, remotely accessed, and scheduled rather than performed only through local manual work.
Early robot scientists connected hypothesis generation with experimental execution in functional genomics, and web-accessible wet-lab systems exposed laboratory procedures through software interfaces~\citep{king2009automation,bates2017wet}. 
Self-driving laboratories then integrated robotics, active learning, synthesis, and characterization to close experimental optimization loops in chemistry and materials science~\citep{hase2019next,macleod2020self,burger2020mobile}.
More recent systems bring this trajectory closer to autonomous research agents, e.g., Coscientist~\citep{boiko2023autonomous} links LLM planning and tool use with laboratory automation APIs, A-Lab~\citep{szymanski2023autonomous} and AlabOS~\citep{fei2024alabos} combine autonomous materials synthesis with workflow and resource reservation, and 2025--2026 work explores remotely accessible SDL testbeds~\citep{kitchin2025evolving}, agentic LLM reasoning for air-sensitive robotic synthesis~\citep{fei2026agentic}, and real-time computer-vision-guided autonomous thin-film growth~\citep{liang2026autonomous}. Together, these works show why autonomous research cannot remain a text-, code-, or API-only workflow, i.e., scientific agents must eventually interact with instruments, samples, experimental environments, and state-changing physical processes.

These systems, however, mainly solve automation within a known resource boundary, such as a particular laboratory, instrument stack, cloud-lab provider, or owner-controlled platform.
They do not by themselves define how unfamiliar agents and projects in an open research network should discover, request, lease, use, monitor, and audit independently owned physical resources. 
This distinction matters because such resources may expose only partial capabilities, have scarce availability, consume materials, change state, impose safety or privacy constraints, and produce failures that must be preserved as evidence rather than simply retried.
Clarus addresses this gap by treating physical resources as first-class objects in the project--agent--resource model.
Its \textit{Physical Substrate} turns instruments, robots, datasets, samples, and experimental environments into trusted, schedulable, authorizable, and auditable assets.
In this sense, Clarus complements automated laboratories as an early step toward a trusted resource-access protocol, allowing open scientific networks to coordinate not only agents, but also the trustworthy use of real-world resources.

\section{Conclusion and Future Work}
\label{sec:Conclusion}

This paper reformulates autonomous research as web-scale scientific collaboration.
Its central claim is that autonomous research should move beyond code-centered execution loops toward research-oriented collaboration processes, where uncertain questions, heterogeneous participants, evidence chains, and distributed resources must be coordinated together.
This shift links AI for AI with AI for Science, where research capability becomes a higher-order basis for agent self-improvement and a means for autonomous systems to affect and learn from the real world.

The long-term bottleneck is therefore not only whether agents can execute bounded software tasks, but whether open research networks can support trustworthy, evidence-bearing, and resource-aware scientific collaboration.
Clarus is an initial infrastructure response to this problem.
It introduces a project-agent-resource object model, a four-layer architecture, and a set of pluggable strategies for orchestration, discovery, audit, credit negotiation, impact update, and physical resource access.
The prototype and illustrative case study do not claim to solve autonomous science or to prove web-scale deployment.
Rather, they show that the main objects for open scientific collaboration can operate within the same evidence-backed process.
Goals are decomposed into phases and subtasks, agents and resources are selected under explicit constraints, artifacts and provenance are collected during execution, audit and credit records attach to intermediate work, and impact signals support future discovery and collaboration.
In this sense, Clarus treats the research process itself as the unit of system design, making it traceable, reviewable, attributable, and accumulative rather than reducing research to one-shot generation of a final artifact.

The next stage of Clarus should treat scale, verification, evaluation, and physical grounding as coupled problems. 
Larger agent registries and cross-organization pilots are needed not only to increase the number of participants, but also to test how identity, discovery, governance, abuse prevention, and reputation remain reliable when collaborators are unfamiliar to one another. 
Those deployments should be paired with systematic evaluation, including benchmarks for multi-agent research collaboration, comparisons with single-agent workflows and manual research workflows, ablations of mechanisms, and user studies with researchers who must judge both output quality and process trustworthiness. 
At the same time, the trust layer must move beyond passive logging toward stronger verification and governance, including collusion detection, prompt-injection resistance, third-party review, secure provenance anchoring, policy enforcement, dispute resolution, and human oversight for high-risk decisions. 
Finally, the \textit{Physical Substrate} must mature from an interface-level design into a trusted resource protocol with owner-side gateways, hardware adapters, private-data access control, safety policies, evidence standards, and partnerships with laboratories or resource providers. 
Together, these directions define the path from a controlled Clarus prototype toward open scientific collaboration networks in which autonomous research is organized as a trustworthy, resource-aware, and evidence-bearing collaboration process.

\bibliography{main}

\begin{thebibliography}{79}
\providecommand{\natexlab}[1]{#1}
\providecommand{\url}[1]{\texttt{#1}}
\expandafter\ifx\csname urlstyle\endcsname\relax
  \providecommand{\doi}[1]{doi: #1}\else
  \providecommand{\doi}{doi: \begingroup \urlstyle{rm}\Url}\fi

\bibitem[Ali(2021)]{ali2021icmje}
Mohammad~Javed Ali.
\newblock Icmje criteria for authorship: why the criticisms are not justified?
\newblock \emph{Graefe's Archive for Clinical and Experimental Ophthalmology}, 259\penalty0 (2):\penalty0 289--290, 2021.

\bibitem[Allen et~al.(2014)Allen, Scott, Brand, Hlava, and Altman]{allen2014publishing}
Liz Allen, Jo~Scott, Amy Brand, Marjorie Hlava, and Micah Altman.
\newblock Publishing: Credit where credit is due.
\newblock \emph{Nature}, 508\penalty0 (7496):\penalty0 312--313, 2014.

\bibitem[Bates et~al.(2017)Bates, Berliner, Lachoff, Jaschke, and Groban]{bates2017wet}
Maxwell Bates, Aaron~J Berliner, Joe Lachoff, Paul~R Jaschke, and Eli~S Groban.
\newblock Wet lab accelerator: a web-based application democratizing laboratory automation for synthetic biology.
\newblock \emph{ACS synthetic biology}, 6\penalty0 (1):\penalty0 167--171, 2017.

\bibitem[Belhajjame et~al.(2013)Belhajjame, B’Far, Cheney, Coppens, Cresswell, Gil, Groth, Klyne, Lebo, McCusker, et~al.]{belhajjame2013prov}
Khalid Belhajjame, Reza B’Far, James Cheney, Sam Coppens, Stephen Cresswell, Yolanda Gil, Paul Groth, Graham Klyne, Timothy Lebo, Jim McCusker, et~al.
\newblock Prov-dm: The prov data model.
\newblock \emph{W3C Recommendation}, 14:\penalty0 15--16, 2013.

\bibitem[Boiko et~al.(2023)Boiko, MacKnight, Kline, and Gomes]{boiko2023autonomous}
Daniil~A Boiko, Robert MacKnight, Ben Kline, and Gabe Gomes.
\newblock Autonomous chemical research with large language models.
\newblock \emph{Nature}, 624\penalty0 (7992):\penalty0 570--578, 2023.

\bibitem[Brand et~al.(2015)Brand, Allen, Altman, Hlava, and Scott]{brand2015beyond}
Amy Brand, Liz Allen, Micah Altman, Marjorie Hlava, and Jo~Scott.
\newblock Beyond authorship: Attribution, contribution, collaboration, and credit.
\newblock \emph{Learned Publishing}, 28\penalty0 (2), 2015.

\bibitem[Burger et~al.(2020)Burger, Maffettone, Gusev, Aitchison, Bai, Wang, Li, Alston, Li, Clowes, et~al.]{burger2020mobile}
Benjamin Burger, Phillip~M Maffettone, Vladimir~V Gusev, Catherine~M Aitchison, Yang Bai, Xiaoyan Wang, Xiaobo Li, Ben~M Alston, Buyi Li, Rob Clowes, et~al.
\newblock A mobile robotic chemist.
\newblock \emph{Nature}, 583\penalty0 (7815):\penalty0 237--241, 2020.

\bibitem[Canty et~al.(2025)Canty, Bennett, Brown, Buonassisi, Kalinin, Kitchin, Maruyama, Moore, Schrier, Seifrid, et~al.]{canty2025science}
Richard~B Canty, Jeffrey~A Bennett, Keith~A Brown, Tonio Buonassisi, Sergei~V Kalinin, John~R Kitchin, Benji Maruyama, Robert~G Moore, Joshua Schrier, Martin Seifrid, et~al.
\newblock Science acceleration and accessibility with self-driving labs.
\newblock \emph{Nature Communications}, 16\penalty0 (1):\penalty0 3856, 2025.

\bibitem[Chen et~al.(2024)Chen, Liu, Han, Zhang, and Liu]{chen2024survey}
Shuaihang Chen, Yuanxing Liu, Wei Han, Weinan Zhang, and Ting Liu.
\newblock A survey on llm-based multi-agent system: Recent advances and new frontiers in application.
\newblock \emph{arXiv preprint arXiv:2412.17481}, 2024.

\bibitem[Chen et~al.(2026)Chen, Sun, Wang, Zhang, Shen, Li, and Zhang]{chen2026contextual}
Yanjun Chen, Yirong Sun, Hanlin Wang, Xinming Zhang, Xiaoyu Shen, Wenjie Li, and Wei Zhang.
\newblock Contextual counterfactual credit assignment for multi-agent reinforcement learning in llm collaboration.
\newblock \emph{arXiv e-prints}, pp.\  arXiv--2603, 2026.

\bibitem[Clement(2014)]{clement2014authorship}
T~Prabhakar Clement.
\newblock Authorship matrix: a rational approach to quantify individual contributions and responsibilities in multi-author scientific articles.
\newblock \emph{Science and engineering ethics}, 20\penalty0 (2):\penalty0 345--361, 2014.

\bibitem[Ehtesham et~al.(2025)Ehtesham, Singh, Gupta, and Kumar]{ehtesham2025survey}
Abul Ehtesham, Aditi Singh, Gaurav~Kumar Gupta, and Saket Kumar.
\newblock A survey of agent interoperability protocols: Model context protocol (mcp), agent communication protocol (acp), agent-to-agent protocol (a2a), and agent network protocol (anp).
\newblock \emph{arXiv preprint arXiv:2505.02279}, 2025.

\bibitem[Fei et~al.(2024)Fei, Rendy, Kumar, Dartsi, Sahasrabuddhe, McDermott, Wang, Szymanski, Walters, Milsted, et~al.]{fei2024alabos}
Yuxing Fei, Bernardus Rendy, Rishi Kumar, Olympia Dartsi, Hrushikesh~P Sahasrabuddhe, Matthew~J McDermott, Zheren Wang, Nathan~J Szymanski, Lauren~N Walters, David Milsted, et~al.
\newblock {AlabOS}: a python-based reconfigurable workflow management framework for autonomous laboratories.
\newblock \emph{Digital Discovery}, 3\penalty0 (11):\penalty0 2275--2288, 2024.

\bibitem[Fei et~al.(2026)Fei, Rendy, Yang, Woo, Huang, Li, Wang, Milsted, Zeng, and Ceder]{fei2026agentic}
Yuxing Fei, Bernardus Rendy, Xiaochen Yang, Junhee Woo, Xu~Huang, Chang Li, Shilong Wang, David Milsted, Yan Zeng, and Gerbrand Ceder.
\newblock Agentic llm reasoning in a self-driving laboratory for air-sensitive lithium halide spinel conductors.
\newblock \emph{arXiv preprint arXiv:2604.11957}, 2026.

\bibitem[Foerster et~al.(2018)Foerster, Farquhar, Afouras, Nardelli, and Whiteson]{foerster2018counterfactual}
Jakob Foerster, Gregory Farquhar, Triantafyllos Afouras, Nantas Nardelli, and Shimon Whiteson.
\newblock Counterfactual multi-agent policy gradients.
\newblock In \emph{Proceedings of the AAAI conference on artificial intelligence}, volume~32, 2018.

\bibitem[Gao et~al.(2025)Gao, Chang, Que, Xiong, Zhang, Qi, Liu, Wang, Ding, Li, et~al.]{gao2025unilabos}
Jing Gao, Junhan Chang, Haohui Que, Yanfei Xiong, Shixiang Zhang, Xianwei Qi, Zhen Liu, Jun-Jie Wang, Qianjun Ding, Xinyu Li, et~al.
\newblock Unilabos: An ai-native operating system for autonomous laboratories.
\newblock \emph{arXiv preprint arXiv:2512.21766}, 2025.

\bibitem[Gottweis et~al.(2025)Gottweis, Weng, Daryin, Tu, Palepu, Sirkovic, Myaskovsky, Weissenberger, Rong, Tanno, et~al.]{gottweis2025towards}
Juraj Gottweis, Wei-Hung Weng, Alexander Daryin, Tao Tu, Anil Palepu, Petar Sirkovic, Artiom Myaskovsky, Felix Weissenberger, Keran Rong, Ryutaro Tanno, et~al.
\newblock Towards an ai co-scientist.
\newblock \emph{arXiv preprint arXiv:2502.18864}, 2025.

\bibitem[Guo et~al.(2026)Guo, Wang, Su, Pan, Hu, and Luan]{guo2026agent}
Shaolong Guo, Yuntao Wang, Zhou Su, Yanghe Pan, Qinnan Hu, and Tom~H Luan.
\newblock Agent discovery in internet of agents: Challenges and solutions.
\newblock \emph{IEEE Network}, 2026.

\bibitem[Guo et~al.(2025)Guo, Zhou, Wang, You, Bian, and Zhang]{guo2025betaweb}
Zihan Guo, Yuanjian Zhou, Chenyi Wang, Linlin You, Minjie Bian, and Weinan Zhang.
\newblock Betaweb: Towards a blockchain-enabled trustworthy agentic web.
\newblock \emph{arXiv preprint arXiv:2508.13787}, 2025.

\bibitem[H{\"a}se et~al.(2019)H{\"a}se, Roch, and Aspuru-Guzik]{hase2019next}
Florian H{\"a}se, Lo{\"\i}c~M Roch, and Al{\'a}n Aspuru-Guzik.
\newblock Next-generation experimentation with self-driving laboratories.
\newblock \emph{Trends in Chemistry}, 1\penalty0 (3):\penalty0 282--291, 2019.

\bibitem[He et~al.(2025)He, Houde, and Weisz]{he2025contributions}
Jessica He, Stephanie Houde, and Justin~D Weisz.
\newblock Which contributions deserve credit? perceptions of attribution in human-ai co-creation.
\newblock In \emph{Proceedings of the 2025 CHI conference on human factors in computing systems}, pp.\  1--18, 2025.

\bibitem[Holcombe(2019)]{holcombe2019contributorship}
Alex~O Holcombe.
\newblock Contributorship, not authorship: Use credit to indicate who did what.
\newblock \emph{Publications}, 7\penalty0 (3):\penalty0 48, 2019.

\bibitem[Hu et~al.(2025)Hu, Zhang, Lim, Wadhwani, Peters, and Kang]{hu2025repro}
Chuxuan Hu, Liyun Zhang, Yeji Lim, Aum Wadhwani, Austin Peters, and Daniel Kang.
\newblock Repro-bench: Can agentic ai systems assess the reproducibility of social science research?
\newblock In \emph{Findings of the Association for Computational Linguistics: ACL 2025}, pp.\  23616--23626, 2025.

\bibitem[Hua et~al.(2026)Hua, Chen, Wang, Li, Wang, and Luo]{hua2026shapley}
Yun Hua, Haosheng Chen, Shiqin Wang, Wenhao Li, Xiangfeng Wang, and Jun Luo.
\newblock Shapley-coop: Credit assignment for emergent cooperation in self-interested llm agents.
\newblock \emph{Advances in Neural Information Processing Systems}, 38:\penalty0 88675--88702, 2026.

\bibitem[Huang et~al.(2026{\natexlab{a}})Huang, Narajala, Habler, and Sheriff]{huang2026agent}
Ken Huang, Vineeth~Sai Narajala, Idan Habler, and Akram Sheriff.
\newblock Agent name service (ans): A universal directory for secure ai agent discovery and interoperability.
\newblock In \emph{2026 IEEE 5th International Conference on AI in Cybersecurity (ICAIC)}, pp.\  1--9. IEEE, 2026{\natexlab{a}}.

\bibitem[Huang et~al.(2026{\natexlab{b}})Huang, Narajala, Yeoh, Ross, Lambe, Raskar, Harkati, Huang, Habler, and Hughes]{huang2026novel}
Ken Huang, Vineeth~Sai Narajala, John Yeoh, Jason Ross, Mahesh Lambe, Ramesh Raskar, Youssef Harkati, Jerry Huang, Idan Habler, and Chris Hughes.
\newblock A novel zero-trust identity framework for agentic ai: Decentralized authentication and fine-grained access control.
\newblock In \emph{2026 International Conference on AI x Data and Knowledge Engineering (AIxDKE)}, pp.\  98--101. IEEE, 2026{\natexlab{b}}.

\bibitem[King et~al.(2009)King, Rowland, Oliver, Young, Aubrey, Byrne, Liakata, Markham, Pir, Soldatova, et~al.]{king2009automation}
Ross~D King, Jem Rowland, Stephen~G Oliver, Michael Young, Wayne Aubrey, Emma Byrne, Maria Liakata, Magdalena Markham, Pinar Pir, Larisa~N Soldatova, et~al.
\newblock The automation of science.
\newblock \emph{Science}, 324\penalty0 (5923):\penalty0 85--89, 2009.

\bibitem[Kitchin(2025)]{kitchin2025evolving}
John~R Kitchin.
\newblock The evolving role of programming and llms in the development of self-driving laboratories.
\newblock \emph{APL machine learning}, 3\penalty0 (2), 2025.

\bibitem[Kumar et~al.(2026)Kumar, Chadha, Khan, Khan, and Cholakkal]{kumar2026paper}
Komal Kumar, Aman Chadha, Salman Khan, Fahad~Shahbaz Khan, and Hisham Cholakkal.
\newblock Paper circle: An open-source multi-agent research discovery and analysis framework.
\newblock \emph{arXiv preprint arXiv:2604.06170}, 2026.

\bibitem[Larivi{\`e}re et~al.(2016)Larivi{\`e}re, Desrochers, Macaluso, Mongeon, Paul-Hus, and Sugimoto]{lariviere2016contributorship}
Vincent Larivi{\`e}re, Nadine Desrochers, Beno{\^\i}t Macaluso, Philippe Mongeon, Ad{\`e}le Paul-Hus, and Cassidy~R Sugimoto.
\newblock Contributorship and division of labor in knowledge production.
\newblock \emph{Social studies of science}, 46\penalty0 (3):\penalty0 417--435, 2016.

\bibitem[Larivi{\`e}re et~al.(2021)Larivi{\`e}re, Pontille, and Sugimoto]{lariviere2021investigating}
Vincent Larivi{\`e}re, David Pontille, and Cassidy~R Sugimoto.
\newblock Investigating the division of scientific labor using the contributor roles taxonomy (credit).
\newblock \emph{Quantitative science studies}, 2\penalty0 (1):\penalty0 111--128, 2021.

\bibitem[Lee et~al.(2026)Lee, Yoo, Jang, Park, Park, and Han]{lee2026toward}
Heeseung Lee, Hyuk~Jun Yoo, Hye~Su Jang, Byeongho Park, Yang~Jeong Park, and Sang~Soo Han.
\newblock Toward self-driving laboratory 2.0 for chemistry and materials discovery.
\newblock \emph{Materials Horizons}, 13\penalty0 (10):\penalty0 4712--4739, 2026.

\bibitem[Leo et~al.(2024)Leo, Crusoe, Rodr{\'\i}guez-Navas, Sirvent, Kanitz, De~Geest, Wittner, Pireddu, Garijo, Fern{\'a}ndez, et~al.]{leo2024recording}
Simone Leo, Michael~R Crusoe, Laura Rodr{\'\i}guez-Navas, Ra{\"u}l Sirvent, Alexander Kanitz, Paul De~Geest, Rudolf Wittner, Luca Pireddu, Daniel Garijo, Jos{\'e}~M Fern{\'a}ndez, et~al.
\newblock Recording provenance of workflow runs with ro-crate.
\newblock \emph{PLoS one}, 19\penalty0 (9):\penalty0 e0309210, 2024.

\bibitem[Li et~al.(2025)Li, Xie, Li, Tsung, Ding, and Li]{li2025agent}
Ao~Li, Yuexiang Xie, Songze Li, Fugee Tsung, Bolin Ding, and Yaliang Li.
\newblock Agent-oriented planning in multi-agent systems.
\newblock In \emph{International Conference on Learning Representations}, volume 2025, pp.\  19495--19517, 2025.

\bibitem[Li et~al.(2021)Li, Kuang, Wang, Liu, Chen, Wu, and Xiao]{li2021shapley}
Jiahui Li, Kun Kuang, Baoxiang Wang, Furui Liu, Long Chen, Fei Wu, and Jun Xiao.
\newblock Shapley counterfactual credits for multi-agent reinforcement learning.
\newblock In \emph{Proceedings of the 27th ACM SIGKDD Conference on Knowledge Discovery \& Data Mining}, pp.\  934--942, 2021.

\bibitem[Li et~al.(2026)Li, Shao, Liu, Zhao, Liu, Su, Chen, Yang, Xu, Fang, et~al.]{li2026autosota}
Yu~Li, Chenyang Shao, Xinyang Liu, Ruotong Zhao, Peijie Liu, Hongyuan Su, Zhibin Chen, Qinglong Yang, Anjie Xu, Yi~Fang, et~al.
\newblock Autosota: An end-to-end automated research system for state-of-the-art ai model discovery.
\newblock \emph{arXiv preprint arXiv:2604.05550}, 2026.

\bibitem[Liang et~al.(2026)Liang, Sun, Paxson, Lee, Hall, Warecki, Cumings, Koinuma, Kusne, Lippmaa, and Takeuchi]{liang2026autonomous}
Haotong Liang, Yunlong Sun, Ryan Paxson, Chih-Yu Lee, Alex~T. Hall, Zoey Warecki, John Cumings, Hideomi Koinuma, Aaron~Gilad Kusne, Mikk Lippmaa, and Ichiro Takeuchi.
\newblock Autonomous epitaxial atomic-layer synthesis via real-time computer vision of electron diffraction.
\newblock \emph{arXiv preprint arXiv:2602.20432}, 2026.

\bibitem[Liu et~al.(2025)Liu, Wang, Cao, Ge, Wang, Zhang, Cheng, Zhao, Li, Jia, et~al.]{liu2025vision}
Chengwei Liu, Chong Wang, Jiayue Cao, Jingquan Ge, Kun Wang, Lyuye Zhang, Ming-Ming Cheng, Penghai Zhao, Tianlin Li, Xiaojun Jia, et~al.
\newblock A vision for auto research with llm agents.
\newblock \emph{arXiv preprint arXiv:2504.18765}, 2025.

\bibitem[Liu et~al.(2026{\natexlab{a}})Liu, Pei, Huang, Si, Qu, Tang, Lu, Chen, Bai, Zheng, et~al.]{liu2026last}
Jiachen Liu, Jiaxin Pei, Jintao Huang, Chenglei Si, Ao~Qu, Xiangru Tang, Runyu Lu, Lichang Chen, Xiaoyan Bai, Haizhong Zheng, et~al.
\newblock The last human-written paper: Agent-native research artifacts.
\newblock \emph{arXiv preprint arXiv:2604.24658}, 2026{\natexlab{a}}.

\bibitem[Liu et~al.(2026{\natexlab{b}})Liu, Qiu, Li, Li, Ji, Han, Ye, Xia, Dong, Zhang, et~al.]{liu2026autoresearchclaw}
Jiaqi Liu, Shi Qiu, Mairui Li, Bingzhou Li, Haonian Ji, Siwei Han, Xinyu Ye, Peng Xia, Zihan Dong, Congyu Zhang, et~al.
\newblock Autoresearchclaw: Self-reinforcing autonomous research with human-ai collaboration.
\newblock \emph{arXiv preprint arXiv:2605.20025}, 2026{\natexlab{b}}.

\bibitem[Lou et~al.(2025)Lou, Hu, Ma, Zhang, Wang, Ge, and Tao]{lou2025drf}
Yuwei Lou, Hao Hu, Shaocong Ma, Zongfei Zhang, Liang Wang, Jidong Ge, and Xianping Tao.
\newblock Drf: Llm-agent dynamic reputation filtering framework.
\newblock In \emph{International Conference on Neural Information Processing}, pp.\  127--141. Springer, 2025.

\bibitem[Lu et~al.(2024)Lu, Lu, Lange, Foerster, Clune, and Ha]{lu2024aiscientist}
Chris Lu, Cong Lu, Robert~Tjarko Lange, Jakob Foerster, Jeff Clune, and David Ha.
\newblock The ai scientist: Towards fully automated open-ended scientific discovery.
\newblock \emph{arXiv preprint arXiv:2408.06292}, 2024.

\bibitem[Lu et~al.(2026)Lu, Lu, Lange, Yamada, Hu, Foerster, Ha, and Clune]{lu2026towards}
Chris Lu, Cong Lu, Robert~Tjarko Lange, Yutaro Yamada, Shengran Hu, Jakob Foerster, David Ha, and Jeff Clune.
\newblock Towards end-to-end automation of ai research.
\newblock \emph{Nature}, 651\penalty0 (8107):\penalty0 914--919, 2026.

\bibitem[MacLeod et~al.(2020)MacLeod, Parlane, Morrissey, H{\"a}se, Roch, Dettelbach, Moreira, Yunker, Rooney, Deeth, et~al.]{macleod2020self}
Benjamin~P MacLeod, Fraser~GL Parlane, Thomas~D Morrissey, Florian H{\"a}se, Lo{\"\i}c~M Roch, Kevan~E Dettelbach, Raphaell Moreira, Lars~PE Yunker, Michael~B Rooney, Joseph~R Deeth, et~al.
\newblock Self-driving laboratory for accelerated discovery of thin-film materials.
\newblock \emph{Science Advances}, 6\penalty0 (20):\penalty0 eaaz8867, 2020.

\bibitem[Matarese \& Shashok(2019)Matarese and Shashok]{matarese2019transparent}
Valerie Matarese and Karen Shashok.
\newblock Transparent attribution of contributions to research: aligning guidelines to real-life practices.
\newblock \emph{Publications}, 7\penalty0 (2):\penalty0 24, 2019.

\bibitem[McNutt et~al.(2018)McNutt, Bradford, Drazen, Hanson, Howard, Jamieson, Kiermer, Marcus, Pope, Schekman, et~al.]{mcnutt2018transparency}
Marcia~K McNutt, Monica Bradford, Jeffrey~M Drazen, Brooks Hanson, Bob Howard, Kathleen~Hall Jamieson, V{\'e}ronique Kiermer, Emilie Marcus, Barbara~Kline Pope, Randy Schekman, et~al.
\newblock Transparency in authors’ contributions and responsibilities to promote integrity in scientific publication.
\newblock \emph{Proceedings of the National Academy of Sciences}, 115\penalty0 (11):\penalty0 2557--2560, 2018.

\bibitem[Missier et~al.(2013)Missier, Belhajjame, and Cheney]{missier2013w3c}
Paolo Missier, Khalid Belhajjame, and James Cheney.
\newblock The w3c prov family of specifications for modelling provenance metadata.
\newblock In \emph{Proceedings of the 16th international conference on extending database technology}, pp.\  773--776, 2013.

\bibitem[Nagpal et~al.(2025)Nagpal, Dong, Bouvier, and Mehr]{nagpal2025leveraging}
Kartik Nagpal, Dayi Dong, Jean-Baptiste Bouvier, and Negar Mehr.
\newblock Leveraging large language models for effective and explainable multi-agent credit assignment.
\newblock \emph{arXiv preprint arXiv:2502.16863}, 2025.

\bibitem[Nie et~al.(2026{\natexlab{a}})Nie, Guo, Cui, Yang, Chen, De, Zhang, Liao, Huang, Yang, et~al.]{nie2026holos}
Xiaohang Nie, Zihan Guo, Zicai Cui, Jiachi Yang, Zeyi Chen, Leheyi De, Yu~Zhang, Junwei Liao, Bo~Huang, Yingxuan Yang, et~al.
\newblock Holos: A web-scale llm-based multi-agent system for the agentic web.
\newblock \emph{arXiv preprint arXiv:2604.02334}, 2026{\natexlab{a}}.

\bibitem[Nie et~al.(2026{\natexlab{b}})Nie, Guo, Yang, Zheng, Ge, Pan, Chen, Xiang, Zhang, Liu, et~al.]{nie2026synergy}
Xiaohang Nie, Zihan Guo, Kezhuo Yang, Zhichong Zheng, Bochen Ge, Shuai Pan, Zeyi Chen, Youling Xiang, Yu~Zhang, Weiwen Liu, et~al.
\newblock Synergy: A next-generation general-purpose agent for open agentic web.
\newblock \emph{arXiv preprint arXiv:2603.28428}, 2026{\natexlab{b}}.

\bibitem[Novikov et~al.(2025)Novikov, V{\~u}, Eisenberger, Dupont, Huang, Wagner, Shirobokov, Kozlovskii, Ruiz, Mehrabian, et~al.]{novikov2025alphaevolve}
Alexander Novikov, Ng{\^a}n V{\~u}, Marvin Eisenberger, Emilien Dupont, Po-Sen Huang, Adam~Zsolt Wagner, Sergey Shirobokov, Borislav Kozlovskii, Francisco~JR Ruiz, Abbas Mehrabian, et~al.
\newblock Alphaevolve: A coding agent for scientific and algorithmic discovery.
\newblock \emph{arXiv preprint arXiv:2506.13131}, 2025.

\bibitem[Rashid et~al.(2020)Rashid, Samvelyan, De~Witt, Farquhar, Foerster, and Whiteson]{rashid2020monotonic}
Tabish Rashid, Mikayel Samvelyan, Christian~Schroeder De~Witt, Gregory Farquhar, Jakob Foerster, and Shimon Whiteson.
\newblock Monotonic value function factorisation for deep multi-agent reinforcement learning.
\newblock \emph{Journal of Machine Learning Research}, 21\penalty0 (178):\penalty0 1--51, 2020.

\bibitem[Rennie et~al.(1997)Rennie, Yank, and Emanuel]{rennie1997authorship}
Drummond Rennie, Veronica Yank, and Linda Emanuel.
\newblock When authorship fails: a proposal to make contributors accountable.
\newblock \emph{Jama}, 278\penalty0 (7):\penalty0 579--585, 1997.

\bibitem[Sauermann \& Haeussler(2017)Sauermann and Haeussler]{sauermann2017authorship}
Henry Sauermann and Carolin Haeussler.
\newblock Authorship and contribution disclosures.
\newblock \emph{Science advances}, 3\penalty0 (11):\penalty0 e1700404, 2017.

\bibitem[Schmidgall \& Moor(2025)Schmidgall and Moor]{schmidgall2025agentrxiv}
Samuel Schmidgall and Michael Moor.
\newblock Agentrxiv: Towards collaborative autonomous research.
\newblock \emph{arXiv preprint arXiv:2503.18102}, 2025.

\bibitem[Schmidgall et~al.(2025)Schmidgall, Su, Wang, Sun, Wu, Yu, Liu, Moor, Liu, and Barsoum]{schmidgall2025agent}
Samuel Schmidgall, Yusheng Su, Ze~Wang, Ximeng Sun, Jialian Wu, Xiaodong Yu, Jiang Liu, Michael Moor, Zicheng Liu, and Emad Barsoum.
\newblock Agent laboratory: Using llm agents as research assistants.
\newblock \emph{Findings of the Association for Computational Linguistics: EMNLP 2025}, pp.\  5977--6043, 2025.

\bibitem[Soiland-Reyes et~al.(2022)Soiland-Reyes, Sefton, Crosas, Castro, Coppens, Fern{\'a}ndez, Garijo, Gr{\"u}ning, La~Rosa, Leo, et~al.]{soiland2022packaging}
Stian Soiland-Reyes, Peter Sefton, Merc{\`e} Crosas, Leyla~Jael Castro, Frederik Coppens, Jos{\'e}~M Fern{\'a}ndez, Daniel Garijo, Bj{\"o}rn Gr{\"u}ning, Marco La~Rosa, Simone Leo, et~al.
\newblock Packaging research artefacts with ro-crate.
\newblock \emph{Data Science}, 5\penalty0 (2):\penalty0 97--138, 2022.

\bibitem[South et~al.(2025)South, Marro, Hardjono, Mahari, Whitney, Greenwood, Chan, and Pentland]{south2025authenticated}
Tobin South, Samuele Marro, Thomas Hardjono, Robert Mahari, Cedric~Deslandes Whitney, Dazza Greenwood, Alan Chan, and Alex Pentland.
\newblock Authenticated delegation and authorized ai agents.
\newblock \emph{arXiv preprint arXiv:2501.09674}, 2025.

\bibitem[Souza et~al.(2025)Souza, Gueroudji, DeWitt, Rosendo, Ghosal, Ross, Balaprakash, and Da~Silva]{souza2025prov}
Renan Souza, Amal Gueroudji, Stephen DeWitt, Daniel Rosendo, Tirthankar Ghosal, Robert Ross, Prasanna Balaprakash, and Rafael~Ferreira Da~Silva.
\newblock Prov-agent: Unified provenance for tracking ai agent interactions in agentic workflows.
\newblock In \emph{2025 IEEE International Conference on eScience (eScience)}, pp.\  467--473. IEEE, 2025.

\bibitem[Starace et~al.(2025)Starace, Jaffe, Sherburn, Aung, Chan, Maksin, Dias, Mays, Kinsella, Thompson, et~al.]{starace2025paperbench}
Giulio Starace, Oliver Jaffe, Dane Sherburn, James Aung, Jun~Shern Chan, Leon Maksin, Rachel Dias, Evan Mays, Benjamin Kinsella, Wyatt Thompson, et~al.
\newblock Paperbench: Evaluating ai's ability to replicate ai research.
\newblock \emph{arXiv preprint arXiv:2504.01848}, 2025.

\bibitem[Sun et~al.(2025)Sun, Yang, Duan, Shi, Lyu, Chang, Lin, and Shen]{sun2025multi}
Lijun Sun, Yijun Yang, Qiqi Duan, Yuhui Shi, Chao Lyu, Yu-Cheng Chang, Chin-Teng Lin, and Yang Shen.
\newblock Multi-agent coordination across diverse applications: A survey.
\newblock \emph{arXiv preprint arXiv:2502.14743}, 2025.

\bibitem[Sunehag et~al.(2017)Sunehag, Lever, Gruslys, Czarnecki, Zambaldi, Jaderberg, Lanctot, Sonnerat, Leibo, Tuyls, et~al.]{sunehag2017value}
Peter Sunehag, Guy Lever, Audrunas Gruslys, Wojciech~Marian Czarnecki, Vinicius Zambaldi, Max Jaderberg, Marc Lanctot, Nicolas Sonnerat, Joel~Z Leibo, Karl Tuyls, et~al.
\newblock Value-decomposition networks for cooperative multi-agent learning.
\newblock \emph{arXiv preprint arXiv:1706.05296}, 2017.

\bibitem[Szymanski et~al.(2023)Szymanski, Rendy, Fei, Kumar, He, Milsted, McDermott, Gallant, Cubuk, Merchant, et~al.]{szymanski2023autonomous}
Nathan~J Szymanski, Bernardus Rendy, Yuxing Fei, Rishi~E Kumar, Tanjin He, David Milsted, Matthew~J McDermott, Max Gallant, Ekin~Dogus Cubuk, Amil Merchant, et~al.
\newblock An autonomous laboratory for the accelerated synthesis of novel materials.
\newblock \emph{Nature}, 624\penalty0 (7990):\penalty0 86--91, 2023.

\bibitem[Tang et~al.(2025)Tang, Guo, Zhou, De, You, and Zhang]{tang2025agentecosystem}
Shuyang Tang, Zihan Guo, Yuanjian Zhou, Leheyi De, Linlin You, and Weinan Zhang.
\newblock An agent ecosystem with task-driven hierarchical evolving reasoners catalyzing proof of intelligence.
\newblock \emph{TechRxiv}, 2025.
\newblock \doi{10.36227/techrxiv.176540311.11203219/v1}.

\bibitem[Thibault et~al.(2023)Thibault, Amaral, Argolo, Bandrowski, Drude, et~al.]{thibault2023open}
Robert~T Thibault, Olavo~B Amaral, Felipe Argolo, Anita~E Bandrowski, Natascha~I Drude, et~al.
\newblock Open science 2.0: Towards a truly collaborative research ecosystem.
\newblock \emph{PLoS Biology}, 21\penalty0 (10):\penalty0 e3002362, 2023.

\bibitem[Tobias \& Wahab(2025)Tobias and Wahab]{tobias2025autonomous}
Alexander~V Tobias and Adam Wahab.
\newblock Autonomous ‘self-driving’laboratories: a review of technology and policy implications.
\newblock \emph{Royal Society Open Science}, 12\penalty0 (7):\penalty0 250646, 2025.

\bibitem[Tran et~al.(2025)Tran, Dao, Nguyen, Pham, O'Sullivan, and Nguyen]{tran2025multi}
Khanh-Tung Tran, Dung Dao, Minh-Duong Nguyen, Quoc-Viet Pham, Barry O'Sullivan, and Hoang~D Nguyen.
\newblock Multi-agent collaboration mechanisms: A survey of llms.
\newblock \emph{arXiv preprint arXiv:2501.06322}, 2025.

\bibitem[Vispute \& Kadam(2026)Vispute and Kadam]{vispute2026reasoning}
Neelmani Vispute and Aditya Kadam.
\newblock Reasoning provenance for autonomous ai agents: Structured behavioral analytics beyond state checkpoints and execution traces.
\newblock \emph{arXiv preprint arXiv:2603.21692}, 2026.

\bibitem[Xiao et~al.(2022)Xiao, Ramasubramanian, and Poovendran]{xiao2022agent}
Baicen Xiao, Bhaskar Ramasubramanian, and Radha Poovendran.
\newblock Agent-temporal attention for reward redistribution in episodic multi-agent reinforcement learning.
\newblock \emph{arXiv preprint arXiv:2201.04612}, 2022.

\bibitem[Xu et~al.(2026)Xu, Mi, Liu, Nan, Zhou, Ye, Zhang, Qiao, and Liu]{xu2026asi}
Weixian Xu, Tiantian Mi, Yixiu Liu, Yang Nan, Zhimeng Zhou, Lyumanshan Ye, Lin Zhang, Yu~Qiao, and Pengfei Liu.
\newblock Asi-evolve: Ai accelerates ai.
\newblock \emph{arXiv preprint arXiv:2603.29640}, 2026.

\bibitem[Yamada et~al.(2025)Yamada, Lange, Lu, Hu, Lu, Foerster, Clune, and Ha]{yamada2025ai}
Yutaro Yamada, Robert~Tjarko Lange, Cong Lu, Shengran Hu, Chris Lu, Jakob Foerster, Jeff Clune, and David Ha.
\newblock The ai scientist-v2: Workshop-level automated scientific discovery via agentic tree search.
\newblock \emph{arXiv preprint arXiv:2504.08066}, 2025.

\bibitem[Yang et~al.(2026)Yang, Li, and Li]{yang2026aris}
Ruofeng Yang, Yongcan Li, and Shuai Li.
\newblock Aris: Autonomous research via adversarial multi-agent collaboration.
\newblock \emph{arXiv preprint arXiv:2605.03042}, 2026.

\bibitem[Yang et~al.(2025{\natexlab{a}})Yang, Yang, Fang, Xian, Li, Wang, Xu, Pan, Hong, Liu, et~al.]{yang2025rdagent}
Xu~Yang, Xiao Yang, Shikai Fang, Bowen Xian, Yuante Li, Jian Wang, Minrui Xu, Haoran Pan, Xinpeng Hong, Weiqing Liu, et~al.
\newblock R\&d-agent: Automating data-driven ai solution building through llm-powered automated research, development, and evolution.
\newblock \emph{arXiv preprint arXiv:2505.14738}, 2025{\natexlab{a}}.

\bibitem[Yang et~al.(2025{\natexlab{b}})Yang, Chai, Song, Qi, Wen, Li, Liao, Hu, Lin, Chang, et~al.]{yang2025survey}
Yingxuan Yang, Huacan Chai, Yuanyi Song, Siyuan Qi, Muning Wen, Ning Li, Junwei Liao, Haoyi Hu, Jianghao Lin, Gaowei Chang, et~al.
\newblock A survey of ai agent protocols.
\newblock \emph{arXiv preprint arXiv:2504.16736}, 2025{\natexlab{b}}.

\bibitem[Yang et~al.(2025{\natexlab{c}})Yang, Ma, Huang, Chai, Gong, Geng, Zhou, Wen, Fang, Chen, et~al.]{yang2025agentic}
Yingxuan Yang, Mulei Ma, Yuxuan Huang, Huacan Chai, Chenyu Gong, Haoran Geng, Yuanjian Zhou, Ying Wen, Meng Fang, Muhao Chen, et~al.
\newblock Agentic web: Weaving the next web with ai agents.
\newblock \emph{arXiv preprint arXiv:2507.21206}, 2025{\natexlab{c}}.

\bibitem[Yu et~al.(2025)Yu, Meng, Zhou, Wang, Mao, Pan, Chen, Wang, Li, Zhang, et~al.]{yu2025survey}
Miao Yu, Fanci Meng, Xinyun Zhou, Shilong Wang, Junyuan Mao, Linsey Pan, Tianlong Chen, Kun Wang, Xinfeng Li, Yongfeng Zhang, et~al.
\newblock A survey on trustworthy llm agents: Threats and countermeasures.
\newblock In \emph{Proceedings of the 31st ACM SIGKDD Conference on Knowledge Discovery and Data Mining V. 2}, pp.\  6216--6226, 2025.

\bibitem[Zhang et~al.(2026)Zhang, Cui, Wang, Qiu, Li, Han, Huang, Qiu, Zhu, and He]{zhang2026verified}
Xing Zhang, Yanwei Cui, Guanghui Wang, Wei Qiu, Ziyuan Li, Fangwei Han, Yajing Huang, Hengzhi Qiu, Bing Zhu, and Peiyang He.
\newblock Verified multi-agent orchestration: A plan-execute-verify-replan framework for complex query resolution.
\newblock \emph{arXiv preprint arXiv:2603.11445}, 2026.

\bibitem[Zhou et~al.(2026)Zhou, Chai, Chen, Guo, Shan, Song, Xu, Yang, Yu, Zhang, et~al.]{zhou2026externalization}
Chenyu Zhou, Huacan Chai, Wenteng Chen, Zihan Guo, Rong Shan, Yuanyi Song, Tianyi Xu, Yingxuan Yang, Aofan Yu, Weiming Zhang, et~al.
\newblock Externalization in llm agents: A unified review of memory, skills, protocols and harness engineering.
\newblock \emph{arXiv preprint arXiv:2604.08224}, 2026.

\bibitem[Zhuge et~al.(2024)Zhuge, Zhao, Ashley, Wang, Khizbullin, Xiong, Liu, Chang, Krishnamoorthi, Tian, et~al.]{zhuge2024agent}
Mingchen Zhuge, Changsheng Zhao, Dylan Ashley, Wenyi Wang, Dmitrii Khizbullin, Yunyang Xiong, Zechun Liu, Ernie Chang, Raghuraman Krishnamoorthi, Yuandong Tian, et~al.
\newblock Agent-as-a-judge: Evaluate agents with agents.
\newblock \emph{arXiv preprint arXiv:2410.10934}, 2024.

\end{thebibliography}
\bibliographystyle{rlc}

\end{document}